\documentclass[a4paper,11pt]{article}
\ifdefined\pdfgentounicode \input{glyphtounicode}\pdfgentounicode=1 \fi
\ifdefined\pdfinterwordspaceon \pdfinterwordspaceon \fi

\usepackage[a4paper,margin=25mm,marginparwidth=16mm,marginparsep=3mm]{geometry}
\usepackage{amsmath,amssymb}
\usepackage{graphicx}
\usepackage{array}
\usepackage{tabularx}
\usepackage{booktabs}
\usepackage{xurl}
\usepackage{xcolor}
\usepackage{hyperref}
\hypersetup{colorlinks=true,linkcolor=black,citecolor=black,urlcolor=blue,
  pdftitle={Where Polanyi Meets Galois: Tacit Knowing as an Algebro-Deterministic Operation, Toward Re-rooting AI-Generated Science},
  pdfauthor={Hiroyuki Chuma, Kanji Otsuka, Yoichi Sato, Kodo Yamauchi}}
\usepackage{bookmark}

\makeatletter
\long\def\@makecaption#1#2{%
  \vskip\abovecaptionskip
  \sbox\@tempboxa{\small #1: #2}%
  \ifdim\wd\@tempboxa>\hsize
    {\small #1: #2\par}%
  \else
    \global\@minipagefalse
    \hb@xt@\hsize{\hfil\box\@tempboxa\hfil}%
  \fi
  \vskip\belowcaptionskip}
\makeatother

\makeatletter
\renewcommand*\l@section[2]{%
  \ifnum \c@tocdepth >\z@
    \addpenalty\@secpenalty
    \addvspace{0.15em \@plus\p@}%
    \setlength\@tempdima{1.5em}%
    \begingroup
      \parindent \z@ \rightskip \@pnumwidth
      \parfillskip -\@pnumwidth
      \leavevmode \bfseries
      \advance\leftskip\@tempdima
      \hskip -\leftskip
      #1\nobreak\hfil
      \nobreak\hb@xt@\@pnumwidth{\hss #2\kern-\p@\kern\p@}\par
    \endgroup
  \fi}
\makeatother

\makeatletter
\def\@maketitle{%
  \newpage\null
  \vspace*{-5.35em}
  \begin{center}%
    \let\footnote\thanks
    {\LARGE \@title \par}%
    \vskip 0.6em
    {\large \lineskip .5em
      \begin{tabular}[t]{c}%
        \@author
      \end{tabular}\par}%
    \vskip 0.4em
    {\large \@date}%
  \end{center}%
  \par\vskip -0.45em}
\makeatother

\makeatletter
\def\@begintheorem#1#2{\trivlist\item[\hskip\labelsep{\bfseries #1\ #2.}]\itshape}
\def\@opargbegintheorem#1#2#3{\trivlist\item[\hskip\labelsep{\bfseries #1\ #2\ (#3).}]\itshape}
\makeatother
\newtheorem{lemma}{Lemma}
\newtheorem{conjecture}{Conjecture}

\newcommand{\CRone}{\ensuremath{\mathrm{CR}_1}}
\newcommand{\CRtwo}{\ensuremath{\mathrm{CR}_2}}

\newcolumntype{L}{>{\raggedright\arraybackslash}X}
\newcolumntype{P}[1]{>{\raggedright\arraybackslash}p{#1}}

\newenvironment{reflist}%
  {\begin{list}{}{\setlength{\leftmargin}{1.6em}\setlength{\itemindent}{-1.6em}%
     \setlength{\itemsep}{3pt}\setlength{\parsep}{0pt}\setlength{\topsep}{4pt}}\small}%
  {\end{list}}

\title{Where Polanyi Meets Galois:\\
Tacit Knowing as an Algebro-Deterministic Operation,\\
Toward Re-rooting AI-Generated Science}
\makeatletter
\let\@fnsymbol\@arabic
\makeatother
\author{Hiroyuki Chuma\thanks{Professor Emeritus, Hitotsubashi University}
  \and Kanji Otsuka\thanks{Professor Emeritus, Meisei University}
  \and Yoichi Sato\thanks{Shuhari System}
  \and Kodo Yamauchi\footnotemark[3]}
\date{November 30, 2026}

\begin{document}
\maketitle
\setcounter{footnote}{3}

\begin{abstract}
We argue that tacit knowing is better modelled as an operating regime of a single memory substrate than as a separate kind of content. Our argument draws on VaCoAl, an algebro-deterministic hyperdimensional memory built from XOR-and-shift operations over $\mathrm{GF}(2)$; on CASRAM, its commercial SRAM-CAM realization; and on PyVaCoAl, a DRAM-CAM realization with the extended VaCoAl architecture.

In VaCoAl/PyVaCoAl, the same directed acyclic graph (DAG) of bound records can be read in two ways. With every address collision repaired (Rescue mode), the trace is exactly recoverable and auditable, but no path outranks another. With collisions tolerated (Don't Care mode), a path-integrated confidence, \CRtwo, emerges; it prunes circuitous routes and ranks direct ones without ever being stated as a proposition. We call this trade-off, first noted in our companion work on Minsky's Emotion Machine, \emph{audit--valuation complementarity} and propose it as a formal counterpart of Polanyi's claim that tacit and explicit knowledge are not sharply divided.

Seven results follow. (1)~Polanyi's from--to structure reads as non-commutative binding and recognition as majority-vote coincidence. (2)~Explicating the valuation completely destroys it (Lemma~\ref{lem:novaluation}), an algebraic analogue of Polanyi's pianist; yet the valuation is articulable at the level of law. Tacitness thus comes in two classes: contingent, which a new query brings out, and regime-bound, which can be stated only in a regime where it no longer exists. Read through them, Polanyi, Nonaka, Tsoukas and Collins each describe part of one substrate. (3)~The same lemma answers a second dualism. The popular moral of dual-process psychology---train slow, deliberate System~2 to tame fast, intuitive System~1---has the shape of the two-store reading of tacit and explicit knowledge that Polanyi rejected: each splits one activity in two, ranks the explicit half higher and sets it over the other. Mercier and Sperber's reply, that reasoning is intuitive inference about reasons, matches Polanyi's, that explicit knowledge is rooted in tacit knowing; in the substrate, complete supervision is full rescue, which erases the valuation it was meant to discipline. (4)~To know a large DAG is to cull it. Simon answered Polanyi's reading of the \emph{Meno} paradox with a test, a generator and statable heuristics. The lineage from AlphaGo to AlphaProof bears out the test---Popperian falsification, automated at the endpoint---while its unstated valuations take over the tacit work of a Kuhnian paradigm, leaving results verified yet rootless. Making their route walkable and aiming falsification upstream when a test fails require a declared regime, a recorded cull, a committed reading and an ordering of what to give up whose fine grades only tolerated imperfection supplies. (5)~Collins's relational, somatic and collective tacit knowledge map onto bound structure, substrate valuation and value binding respectively; the last is absent from the substrate. (6)~Our mentor--student DAG is a graph of tacit-knowledge transmission. (7)~Terence Tao's observation that AI-solved problems now spawn few follow-up activities is diagnosed as \emph{rootless explicit knowledge}: results that no human holds tacitly, in art, craft and skill as in mathematics. We propose a division of labour that re-roots them: LLMs generate, VaCoAl/PyVaCoAl keeps the auditable trace and its confidence, and the human community values, with intellectual virtues that cannot be offloaded. We close with testable predictions.
\end{abstract}

\clearpage
\begingroup\small\linespread{0.9}\selectfont
\tableofcontents
\endgroup
\clearpage

\section{Introduction}\label{sec:intro}

Polanyi opened his account of ``tacit knowing'' with a face (Polanyi 1966, 4). We can pick out a familiar face among a thousand, even a million, yet we usually cannot say how we do it. The police Identikit lets a witness compose a likeness from a stock of drawn noses, mouths and hairlines, but only because the witness can tacitly match each drawn feature against the remembered whole---which is why, in his phrase, ``we can know more than we can tell'' (Polanyi 1966, 4\nobreakdash--5).

The first commercial deployment of the VaCoAl architecture is also a face. Shuhari System's CASRAM, which realizes VaCoAl on SRAM inside an FPGA, is offered in a webcam face-recognition system (\href{https://shuharisystem.com/?page_id=1336}{Shuhari System n.d.}): a GPU turns each face into an embedding no one can interpret, and CASRAM matches it by block-wise majority vote, whose agreement can be read off as a confidence. The system knows more than it can tell. Unlike us, it can also show how strongly its parts agreed. This paper takes that coincidence seriously.

The question has become urgent for a reason outside our programme. In July 2026 a counterexample to the Jacobian conjecture in dimension three, found with the Fable AI model, was announced; it can be verified by a brief calculation. Terence Tao observed that the construction looks like a ``massive miracle'', and wrote a ``digestion'' of it (\href{https://terrytao.wordpress.com/2026/07/21/a-digestion-of-the-jacobian-conjecture-counterexample/}{Tao 2026b}). In a Mathstodon thread he described the broader pattern (\href{https://mathstodon.xyz/@tao}{Tao 2026a}): results come from prompters who cannot answer questions about them or give talks, fewer seminars, workshops and collaborations follow, few newcomers join the field, and open directions are withheld for fear of being scooped. Grant Sanderson compressed the point into one line: ``every AI-generated proof is born an unsolved exposition problem'' (\href{https://terrytao.wordpress.com/2026/09/18/if-math-is-more-than-proof-we-need-to-better-celebrate-the-rest-of-it/}{Sanderson 2026}).

We read these observations through Polanyi's dictum that ``all knowledge is either tacit or rooted in tacit knowledge'' (Polanyi 1969, 144). An AI-produced result that no human holds tacitly is explicit knowledge without a root. It can be checked but is hard to teach, extend or value, because those activities draw on tacit knowing. We call such results \emph{rootless explicit knowledge}, and we ask what kind of substrate could help re-root them. The pattern is not confined to mathematics: Zoran argues that generative systems ``make it possible to produce forms of a sophistication that previously required the maker's becoming, without that becoming'' (Zoran 2026). Experienced endoscopists detected fewer adenomas without AI once AI assistance had become routine (Budzy\'{n} et al.\ 2025), and developers who learned a new library with AI assistance understood it less well than those who learned it without (\href{https://arxiv.org/abs/2601.20245}{Shen \& Tamkin 2026}).

{A second relation has the same structure.} The moral commonly drawn from Kahneman's \emph{Thinking, Fast and Slow} (Kahneman 2011) is that errors come from a fast, intuitive System~1 and corrections from a slow, deliberate System~2, so the second should be trained to tame the first. Mercier and Sperber reply that there are not two systems to rank: reasoning is itself intuitive inference, applied to reasons (Mercier \& Sperber 2017). We argue that this moral has the same shape as the popular reading of Polanyi, which sorts knowledge into a tacit and an explicit store and sets the second to capture the first, and that Polanyi's objection to that reading is Mercier and Sperber's objection in another field: the explicit, deliberate part is something the tacit, intuitive part does.

Our answer builds on the VaCoAl programme (Chuma, Otsuka \& Sato 2026a--g). VaCoAl is a hyperdimensional memory whose every operation reduces to XOR-and-shift over $\mathrm{GF}(2)$. Our companion paper on Minsky's Emotion Machine observed that audit and valuation are one circuit at opposite settings of one parameter (2026f): the same substrate yields either an exactly auditable trace or an emergent valuation of paths, not both at full strength. The step taken here is epistemological: we argue that this trade-off is the right formal model of the tacit--explicit relation, and of the relation between intuition and deliberate reasoning.

A third line of work presses the same question from the side of the machines. Polanyi used the paradox of Plato's \emph{Meno} to argue that seeing a problem already takes a tacit foreknowledge of what will solve it (Polanyi 1966, 21\nobreakdash--24). Simon replied that nothing tacit is needed: a problem is well posed once there is a test that recognizes a solution and a generator that produces candidates, guided by rules of thumb that can be stated (Simon 1976; Newell \& Simon 1976). The lineage from AlphaGo to AlphaProof has built Simon's reply at scale: a learned network proposes and an exact checker disposes (Silver et al.\ 2016; Trinh et al.\ 2024; Hubert et al.\ 2025; \href{https://arxiv.org/abs/2605.22763}{Tsoukalas et al.\ 2026}). Yet the follow-up gap does not close where the test is strongest: a formally verified proof is the purest case of a result that can be checked and that no one yet understands. A large DAG holds far more than can ever be drawn out of it, so every searcher has to cull. A checker audits an endpoint and values no path, and the valuation that steers the search stays in learned weights: the test has moved the tacit component, not removed it (\S\ref{sec:culls}).

The paper makes five contributions, which gather the seven results listed in the abstract.
\begin{enumerate}
\item \textbf{An algebraic reading of Polanyi's structure of tacit knowing (\S\ref{sec:polanyi}).} The from--to relation becomes non-commutative binding, subsidiary awareness becomes bundled-but-unqueried content, focal awareness becomes unbinding, the Identikit becomes role--filler composition, and recognition becomes majority-vote coincidence.
\item \textbf{Audit--valuation complementarity (\S\ref{sec:complementarity}).} Repairing every collision makes the trace exact and erases path valuation; tolerating collisions creates valuation and makes the read approximate. Polanyi's pianist, who falters when he attends to his fingers (Polanyi 1958, 56), has an exact algebraic counterpart. The trade-off separates contingently tacit structure, which a new query brings out, from regime-bound tacitness, which explication destroys. Read through these two classes, Polanyi, Nonaka, Tsoukas and Collins each describe part of one substrate correctly.
\item \textbf{A second dualism under the same dial (\S\ref{sec:twodualisms}).} The taming moral drawn from Kahneman's two systems shares its split, its ranking and its prescription of supervision with the two-store reading of knowledge. Mercier and Sperber's reply to the first and Polanyi's reply to the second coincide part for part, and in the substrate complete supervision is the regime in which Lemma~\ref{lem:novaluation} erases the valuation.
\item \textbf{Three ways to cull a DAG (\S\ref{sec:culls}, \S\ref{sec:verifier}, \S\ref{sec:falsification}).} We compare the approval-shaped cull of language models tuned by human preference, the verifier-shaped cull of the lineage from AlphaGo to AlphaProof Nexus, and the substrate-born cull of \CRtwo. Restated in Popper's and Kuhn's vocabularies, the checker keeps falsification at the endpoint while the cull takes over the functions of a paradigm, and the normal product is a result that is verified but rootless. To aim a failed test upstream, a committed reading has to show where the test bears and an ordering has to say what to give up first. What can be stated, such as provenance, orders retraction only coarsely; a finer ordering exists only where imperfection is tolerated.
\item \textbf{A diagnosis and a remedy for the follow-up gap (\S\S\ref{sec:dag}--\ref{sec:reroot}).} The mentor--student DAG is reread as a graph of tacit-knowledge transmission (\S\ref{sec:dag}). Rootless explicit knowledge is defined and extended from mathematics to art, craft and skilled work (\S\ref{sec:rootless}). A three-layer division of labour---LLM generation, VaCoAl/PyVaCoAl trace and confidence, community valuation---is proposed to re-root AI results, and virtue epistemology explains why the third layer stays human (\S\ref{sec:reroot}).
\end{enumerate}

Section~\ref{sec:predictions} states testable predictions, limitations and open problems. We use ``tacit'' in the structural sense defined in \S\ref{sec:complementarity}: information that shapes retrieval and inference without being represented as a proposition the system can state. Whether a memory substrate can be a knower in Polanyi's ``personal'' sense (Polanyi 1958) lies outside our claims. Our claim is narrower and testable. The structure Polanyi described---an integration that focal scrutiny disrupts, yet which can be articulated at another level---appears in VaCoAl/PyVaCoAl as an exact, parameter-controlled trade-off.

\section{Background}\label{sec:background}

Four literatures meet in this paper: the debate on whether tacit knowledge can be made explicit, the VaCoAl programme, the dual-process debate in the psychology of reasoning, and the exchange between Polanyi and Simon over the paradox of the \emph{Meno} and the machines since built on Simon's side of it.

\subsection{Five positions on tacit knowledge}\label{sec:positions}

Polanyi described tacit knowing as a from--to relation: we attend \emph{from} subsidiary particulars \emph{to} a focal whole, and the knower's participation integrates the two (Polanyi 1958; 1966). He insisted that tacit and explicit knowledge are not sharply divided (Polanyi 1969). Kuhn located much of a scientist's competence in shared exemplars acquired through practice (Kuhn 1962/1970). Nonaka and Takeuchi treated tacit and explicit knowledge as distinct contents that convert into each other in a spiral of four modes: socialization (tacit to tacit, by shared practice), externalization (tacit to explicit, as metaphor, analogy, concept, hypothesis or model), combination (explicit to explicit) and internalization (explicit to tacit, by doing) (Nonaka \& Takeuchi 1995). Tsoukas objected that conversion ignores the ineffability Polanyi stressed (Tsoukas 2003), and Nonaka and von Krogh replied that the two lie on one continuum, at whose tacit end embodied knowledge cannot be represented (2009). Collins showed that the word covers three different things: relational, somatic and collective tacit knowledge (Collins 2010).

\begin{table}[tbp]
\centering\small
\caption{Five positions on tacit knowledge.}\label{tab:positions}
\smallskip
\begin{tabularx}{\linewidth}{P{0.21\linewidth}LL}
\toprule
Position & What makes knowledge tacit & Can it become explicit? \\
\midrule
Polanyi (1958, 1966, 1969) & Subsidiary particulars integrated toward a focal whole by the knower & Not sharply divided; explicit knowledge always rests on tacit understanding \\
Kuhn (1962/1970) & Competence carried by shared exemplars learned in practice & Only partly; exemplars do work that rules cannot \\
Nonaka \& Takeuchi (1995); Nonaka \& von Krogh (2009) & Personal, context-bound know-how and mental models; at the far end, embodied knowledge & Yes, by externalization through metaphor, analogy, concept, hypothesis, model; the embodied end cannot be represented (2009) \\
Tsoukas (2003) & An ineffable dimension of skilled knowing & No; it can be displayed and re-punctuated in dialogue, not converted \\
Collins (2010) & Three kinds: relational, somatic, collective & Relational: yes. Somatic: by machine reproduction. Collective: not in the foreseeable future \\
\bottomrule
\end{tabularx}
\end{table}

Table~\ref{tab:positions} shows why the debate has stalled: each position answers a different question. We propose that VaCoAl/PyVaCoAl is a single formal object in which all four questions have answers (\S\ref{sec:fourpositions}). One further reading, held by none of the five, is the most widely circulated: tacit and explicit knowledge are two stores, the explicit one ranks higher, and the task is to move as much as possible into it. Polanyi's dictum rules this picture out, and Nonaka and von Krogh disown it (2009). Section~\ref{sec:twodualisms} argues that it has a twin in the psychology of reasoning.

\subsection{The VaCoAl programme}\label{sec:programme}

The programme consists of seven arXiv papers, cited here as 2026a--g in order of submission (Table~\ref{tab:papers}); numbers and section references for 2026a follow its version~5 (21 May 2026).

\begin{table}[tbp]
\centering\small
\caption{The seven papers of the VaCoAl programme.}\label{tab:papers}
\smallskip
\begin{tabularx}{\linewidth}{P{0.08\linewidth}P{0.38\linewidth}L}
\toprule
Ref. & Paper & What this paper takes from it \\
\midrule
2026a & \href{https://arxiv.org/abs/2604.11665}{Beyond LLMs, Sparse Distributed Memory, and Neuromorphics: VaCoAl} (arXiv:2604.11665) & Architecture, Galois-field diffusion, Rescue vs Don't Care, emergent STDP-like selection, mathematical auditability, HDC-AI as a complement to LLMs \\
2026b & \href{https://arxiv.org/abs/2605.15652}{Bridging Silicon and the Hippocampus} (arXiv:2605.15652) & Two hippocampal regimes, \CRtwo{} as a model of multiplicative replay decay, Pearl's rungs 2 and 3 \\
2026c & \href{https://arxiv.org/abs/2605.21379}{How to Build Marcus's Algebraic Mind: Substrate over Galois Fields} (arXiv:2605.21379) & Operations over variables, recursive structure, individuals vs kinds; reversible binding and non-commutative bundling \\
2026d & \href{https://arxiv.org/html/2605.31470}{Institutions and the Transmission of Upper-Tail Human Capital} (arXiv:2605.31470) & Leibnizian hourglass, 46-fold rise in society membership, possession vs transmission, the Monastery Wall \\
2026e & \href{https://arxiv.org/pdf/2607.16573}{How to Build Marcus's Algebraic Mind: Thagard's Brain--Mind Viewpoint} (arXiv:2607.16573) & A reversible substrate for deeply nested bindings; demonstrated vs physically realized claims \\
2026f & \href{https://arxiv.org/abs/2607.16623}{How to Build Marcus's Algebraic Mind: Minsky's Emotion-Machine Viewpoint} (arXiv:2607.16623) & Introspection as exact recovery of one's own trace; valuation in Don't Care, audit in Rescue \\
2026g & \href{https://www.alphaxiv.org/abs/2608.17915}{Computational Prosopography across a Millennium} (arXiv:2608.17915) & Tool criticism: every ranking decision is reconstructable, and the macro-structures survive switching the bias off \\
\bottomrule
\end{tabularx}
\end{table}

VaCoAl (Vague Coincident Algorithm) replaces the random projection of Kanerva's sparse distributed memory (Kanerva 1988) with deterministic diffusion over $\mathrm{GF}(2)$ (2026a). An input polynomial $P(x)$ is partitioned into $N$ blocks; each block $b$ diffuses its segment with its own long-period feedback polynomial $G_b(x)$ and writes an entry address into its own SRAM or DRAM table:
\begin{equation}
\Psi_b\bigl(P(x)\bigr) = x^{m} P(x) \bmod G_b(x).
\end{equation}
Retrieval is a majority vote over the $N$ blocks, and the share of agreeing blocks, \CRone, is a confidence read off every retrieval. Binding is XOR-and-shift. With $\sigma$ a fixed shift, a role $R$ is bound to a filler $F$ and recovered exactly in the noiseless case (2026a; 2026c):
\begin{equation}
\mathrm{Bind}(R,F) = R \oplus \sigma(F), \qquad
\mathrm{Unbind}\bigl(\mathrm{Bind}(R,F),\,R\bigr) = \sigma^{-1}\bigl(R \oplus R \oplus \sigma(F)\bigr) = F.
\end{equation}
Because $\sigma$ acts on the second argument only, $\mathrm{Bind}(R,F)$ differs from $\mathrm{Bind}(F,R)$ (2026c). Multi-hop traversal multiplies per-step confidences into a path integral:
\begin{equation}
\mathrm{CR}_2(n) = \mathrm{CR}_2(n-1)\cdot \mathrm{CR}_1(n-1) \;\approx\; \mathrm{CR}_1^{\,n}.
\end{equation}

Two operating modes matter here. In Rescue mode a four-stage circuit repairs every block collision; with full rescue the output is bitwise identical to a Python dictionary, \CRone{} is pinned at 1.0, and candidates inside the frontier fall back to lexicographic order. In Don't Care mode collided blocks abstain, and the residual variance in \CRone{} accumulates in \CRtwo{}, the only basis for ranking candidates (2026a). The results used here are demonstrated in the software version of PyVaCoAl; none depends on the speed and power of CASRAM (2026e).

\subsection{Tacit knowledge and machines}\label{sec:machines}

Whether machines can hold tacit knowledge is an older question than this programme: Dreyfus argued that symbolic AI would fail because skilled performance rests on embodied, situated know-how that rules cannot capture (Dreyfus 1972), and language models have reopened the question. Those debates ask whether a given kind of machine, usually a learned network whose internals are hard to read, has tacit knowledge. We ask where, in a substrate whose every operation can be inspected, the tacit--explicit boundary falls, and what moves it.

\subsection{Dual-process theories and the argumentative theory of reason}\label{sec:dualprocess}

Dual-process theories distinguish rapid, autonomous Type~1 processes from slower Type~2 processes that load on working memory (Evans \& Stanovich 2013). Kahneman gave the distinction its best-known names, System~1 and System~2 (Kahneman 2011). His own presentation is guarded, but what circulates is a simpler moral: intuition is where errors come from, deliberation is where they are corrected, and the task is to train the second to tame the first. Evans and Stanovich list the equation of Type~1 with bad thinking among the fallacies about their own theory (2013). We call it the \emph{taming moral}; it, and not the theories from which it is drawn, is our target.

Mercier and Sperber reject the core assumption of the dual-process picture, that intuition and reasoning are the work of two different kinds of mechanism (Mercier \& Sperber 2017, ch.~9). Inference is what all cognition does, in other animals as in humans, and nearly all of it is intuitive (ch.~3). Reason is one inferential module, specialized for reasons: ``reasoning is a use of intuitive inferences about reasons'' (ch.~9). The biases are reasoning's own: the myside bias is a feature of reasoning when it produces reasons (ch.~11). And reasons are for others: reasoning serves to justify and convince and to evaluate others' arguments, so it does well in exchange and poorly in solitude (ch.~10).

\subsection{Search, tests and the \emph{Meno} paradox}\label{sec:meno}

Polanyi's best-known argument that discovery needs tacit knowing starts from a paradox in Plato's \emph{Meno}. Someone who knows the answer has nothing to seek, and someone who does not would not recognize it on meeting it. Polanyi's way out is that the searcher knows tacitly: to see a problem is already to have ``an intimation of the coherence of hitherto not comprehended particulars'' (Polanyi 1966, 21), a foreknowledge that cannot be stated and that tells a scientist which problems are good and guides the search for their solution (1966, 21\nobreakdash--24).

Simon replied that nothing tacit is needed (Simon 1976). One can know exactly what would count as a solution without having one: a problem is well posed when there is a test to which any proposed solution can be put and a generator that produces candidates, as for Goldbach's conjecture. Three years earlier he had argued, against Popper, that scientific discovery does have a logic (Simon 1973). In the same year Newell and Simon added that to solve a problem intelligently is to avoid the exponential explosion of search by information drawn from the task itself, and that there is ``no mystery where the information that guided the search came from'' (Newell \& Simon 1976). Simon's reply therefore has two halves: a test and a generator suffice to pose a problem and to recognize its solution, and what guides the generator through a space too large to survey is heuristic information that can, in principle, be stated.

The first half has since been built at scale. In AlphaGo, learned networks supply the judgements and the rules of Go settle who has won (\href{https://doi.org/10.1038/nature16961}{Silver et al.\ 2016}); in AlphaGeometry a language model proposes constructions and a symbolic engine carries out every deduction (\href{https://doi.org/10.1038/s41586-023-06747-5}{Trinh et al.\ 2024}); AlphaProof learns by reinforcement inside the Lean proof assistant, which decides whether a proof is a proof (\href{https://doi.org/10.1038/s41586-025-09833-y}{Hubert et al.\ 2025}). AlphaProof Nexus, whose language models edit Lean proof sketches checked by the compiler, has proved open Erd\H{o}s problems (\href{https://arxiv.org/abs/2605.22763}{Tsoukalas et al.\ 2026}). Each pairs a learned generator with an exact test. We call these systems the \emph{verifier lineage}.

The lineage settles part of the old dispute. Simon was right that a test and a generator are enough to produce solutions that nobody had found. The second half of his reply has fared differently. The strongest members of the lineage guide their search not by heuristics that could be written down, as the Logic Theorist's were, but by valuations learned in networks that give no reasons. The lineage thus leads back to what Polanyi was pointing at: what the generator's sense of where to look consists of, who decides which problems are given a test at all, and what a result still lacks once the test has passed it (\S\S\ref{sec:simon}, \ref{sec:culls}, \ref{sec:verifier}).

\section{Polanyi's structure of tacit knowing, read algebraically}\label{sec:polanyi}

Each element of Polanyi's anatomy of tacit knowing (Polanyi 1958; 1966) has a counterpart among VaCoAl's operations, and the correspondence is exact enough to tabulate. Table~\ref{tab:dictionary} is a dictionary, not yet an argument; the argument begins in \S\ref{sec:complementarity}, where the dictionary is used to locate the tacit--explicit boundary.

\begin{table}[tbp]
\centering\small
\caption{Polanyi's anatomy of tacit knowing and its counterparts in VaCoAl.}\label{tab:dictionary}
\smallskip
\begin{tabularx}{\linewidth}{P{0.2\linewidth}LL}
\toprule
Polanyi's term & What it names & VaCoAl counterpart \\
\midrule
From--to relation & We attend from proximal particulars to a distal whole; the relation has a direction & Non-commutative binding $\mathrm{Bind}(R,F)=R\oplus\sigma(F)$; directed DAG edges \\
Subsidiary awareness & Particulars present and functioning but not attended & Fillers bundled into a composite vector that no query has unbound \\
Focal awareness & The object of attention & The filler returned by unbinding the composite against a chosen role \\
Tacit integration & Particulars fuse into a meaningful whole & Bundling and majority vote: the whole is the argmax of many partial agreements \\
The Identikit & A face communicated by assembling chosen features & Role--filler composition, $\mathrm{Face}=\sum_i \mathrm{Bind}(\mathrm{role}_i,\mathrm{feature}_i)$ \\
Knowing how without telling how & Recognition without stateable rules & Coincidence of votes across blocks: \CRone{} reports how much agreement, not why \\
Indwelling & We dwell in the subsidiaries, as in a probe or a tool & Fixed diffusion polynomials, used on every read and never inspected during use \\
\bottomrule
\end{tabularx}
\end{table}

\subsection{From--to as directed binding}\label{sec:fromto}

Polanyi's relation is asymmetric: the ``proximal'' term is attended \emph{from}, the ``distal'' term \emph{to} (Polanyi 1966, 10), and reversing them changes what is known. Binding in VaCoAl has the same asymmetry, because the shift $\sigma$ marks which argument is the filler; $\mathrm{Bind}(R,F)$ and $\mathrm{Bind}(F,R)$ are different vectors (2026c). The prosopographic graph carries the asymmetry in its data: its edges run from mentor to student, and 369 bidirectional assertions among roughly 470,000 records had to be removed before it could be traversed as a DAG (\href{https://www.alphaxiv.org/abs/2608.17915}{2026g}).

\subsection{Subsidiary and focal as two reads of one vector}\label{sec:reads}

A composite $H=\sum_i \mathrm{Bind}(R_i,F_i)$ carries every filler at once, and none of them is singled out until a query asks for it. Unbinding $H$ against a role $R_k$ brings $F_k$ forward while the other terms recede into noise near the 0.5 similarity baseline. The programme uses both kinds of read on the same vectors: the ``calculus degree'' of a scholar is a holistic read, the Hamming similarity of a whole composite to a bundled reference concept, and extracting a scholar's FIELD component is an analytic read (2026a; \href{https://arxiv.org/html/2605.31470}{2026d}). Polanyi's two kinds of awareness therefore become two kinds of read on one representation. The holistic read answers ``how much is this like that?''\ without naming parts; the analytic read names one part and loses the whole. Polanyi observed that one cannot attend focally to a particular while keeping it in its subsidiary role. In the algebra, the integrated similarity is not recoverable from any single unbound filler.

\subsection{The Identikit as role--filler composition}\label{sec:identikit}

Polanyi used the police Identikit to show that communicating a face still depends on tacit matching: the witness can assemble a likeness from drawn features only because she can match each feature against the remembered face, and she cannot say how she does that (Polanyi 1966). Hyperdimensional computing gives the composition an exact form: a face record is a bundle of bound pairs, $\sum_i \mathrm{Bind}(\mathrm{role}_i,\mathrm{feature}_i)$, each recoverable by unbinding its role. In CASRAM's face-recognition system, by contrast, a GPU extracts an embedding from each face and CASRAM performs the match by block-wise vote; we do not claim that it binds faces as role--filler pairs.

What the product shows is that the tacit--explicit boundary can be moved. For Polanyi's witness, the composition could be told and the matching could not. In this system it is the other way round. The matching is done by a memory whose decisions are majority votes with a readable confidence (2026a), while the encoding that produced the embedding is the part no one can state. The boundary has moved, and now runs along a physical link between two pieces of hardware. That is the thesis of \S\ref{sec:complementarity} in miniature: where the boundary falls depends on the substrate and its operations, not on the knowledge itself.

\subsection{Knowing how without telling how}\label{sec:telling}

The name VaCoAl, Vague Coincident Algorithm, already states the model of recognition: an answer is what many partial, individually unreliable votes coincide on (2026a). Such a system cannot state why the winner won, but it can state how much its blocks agreed and how the agreement decayed along a path. This is a third condition between Polanyi's poles. The system is articulate about agreement and inarticulate about reasons. People are usually inarticulate about both; we feel more or less sure but cannot count our votes. The partial telling matters for \S\ref{sec:reroot}, because confidence and trace are exactly what a human needs in order to re-root a result.

\subsection{Indwelling and the fixed scaffold}\label{sec:indwelling}

Polanyi described tools and probes as things we come to ``dwell in'' (Polanyi 1966, 17\nobreakdash--18): the blind man feels the end of his stick, not its pressure on his palm. VaCoAl's per-block feedback polynomials play this role. They are used on every read and write and shape everything the system can distinguish, yet are never themselves queried. The programme links this design choice to biology: grid-cell codes appear to reach hippocampal CA3 through a fixed projection laid down early in life, onto which later memories are orthogonalized (2026a; 2026d). On this reading indwelling is not a mystery added to cognition. It is the ordinary condition of any substrate whose orthogonalizer is hardwired and therefore invisible to its own queries.

\section{Audit--valuation complementarity: tacit knowing as an operating regime}\label{sec:complementarity}

In VaCoAl/PyVaCoAl the boundary between tacit and explicit is not a property of content. It is set by the operating regime and by the family of read operations available, and the regimes trade exact auditability against emergent valuation.

\subsection{A relational definition}\label{sec:def}

Let $M$ be a memory in operating regime $\rho$ (for VaCoAl/PyVaCoAl: the rescue rate RR, memory depth $2^m$, block count $N$ and frontier size FS), and let $Q$ be the family of read operations that can be applied to it. We say that information $I$ is:
\begin{itemize}
\item \textbf{explicit} in $(M,\rho,Q)$ if some query in $Q$ returns $I$ as a stated item;
\item \textbf{structurally tacit} in $(M,\rho,Q)$ if $I$ changes the output of some query in $Q$, but no query in $Q$ returns $I$ as an item.
\end{itemize}
Structurally tacit information then comes in two kinds:
\begin{itemize}
\item \textbf{contingently tacit} if some enlarged family $Q'$, run under the same regime $\rho$, returns $I$ as an item;
\item \textbf{regime-bound tacit} if every family that returns $I$ as an item requires a regime in which $I$ no longer exists.
\end{itemize}

In plain words, a \emph{query} is a question put to the memory, an \emph{item} is an answer that states something outright, as a lookup returns a stored record, and the \emph{regime} is the way the memory is being run.\footnote{We use `regime' for the operating setting $\rho$ of a memory. What a scientific community shares when it agrees on which problems to pose and how to test answers, the regime Kuhn called a paradigm, is a regime of another kind (\S\ref{sec:falsification}).} In VaCoAl/PyVaCoAl the setting that matters most here is the rescue rate RR, which decides whether address collisions are repaired (Rescue mode) or tolerated (Don't Care mode). Board games serve as running examples (sources in \S\ref{sec:rootless-def}): a game program is the memory, the way it plays its regime, and what we can ask it its queries.

\emph{Structurally tacit.} Information can be at work in a memory without ever being stated by it: the answers come out differently with it than without it, yet none of the questions we can ask gets it back as an answer. The knowledge that lets us recognize a face decides the answer, ``that is my friend'', but it is not part of the answer. Its tacitness is a fact about the memory and its queries, not about the information. Suppose the only question we can put to AlphaGo is ``what is your move?'' Everything that led it to Move~37, its famous move against Lee Sedol in 2016, is then structurally tacit. The futures it weighed, and the hunches it weighed them with, made it choose that move, yet the answer itself is a single stone.

\emph{Contingently tacit.} Some structurally tacit information is unstated only because no one has yet asked the right question. Add a new kind of question, change nothing else about how the memory is run, and the information comes out as an answer. AlphaGo's search kept the futures it weighed in a game tree. Add the question ``show me your tree'', and those futures come out as stated items, while AlphaGo goes on playing exactly as before. The reasons of IBM's chess program Deep Blue were of the same kind: rules that people had written.

\emph{Regime-bound tacit.} Other structurally tacit information exists only while the memory is run in a particular way. A question that could state it works only if the memory is run differently, and once the memory is run differently, it no longer contains the information. Asking for it destroys it. The human case is Polanyi's pianist (\S\ref{sec:pianist}): if he shifts his attention to his fingers to examine his skill, the skill is no longer there to be examined. AlphaGo's hunches give a board-game picture: demand that every judgment rest on a rule a person can read, as Deep Blue's did, and AlphaGo would lose the sense of who is ahead at a glance that it needed in order to win (\href{https://www.technologyreview.com/2026/10/02/1145639/dont-be-fooled-llms-dont-reason/}{Graepel 2026}). This is an illustration, not a finding. In VaCoAl/PyVaCoAl the corresponding case is exact (\S\ref{sec:tradeoff}). The traversal's preference for direct over circuitous lineages guides the search while collisions are tolerated, and vanishes when every collision is repaired.

The definition makes tacitness relative to a substrate and its operations, as Collins's relational tacit knowledge is relative to social arrangements. It also turns a recurring question---can a new operation produce explicit knowledge?---into a question about enlarging $Q$. Table~\ref{tab:classes} fills the classes from the programme's own findings.

\begin{table}[tbp]
\centering\small
\caption{The classes of \S\ref{sec:def} filled with findings of the programme.}\label{tab:classes}
\smallskip
\begin{tabularx}{\linewidth}{P{0.17\linewidth}LLP{0.17\linewidth}}
\toprule
Class & Example from the programme & What exposes it & Under full rescue \\
\midrule
Explicit & A single recorded assertion that one person trained another & Direct lookup & Unchanged \\
Contingently tacit & The Leibnizian hourglass: 5.3 paths per node upstream, 53.4 downstream & Exhaustive path counting over 25.5 million paths & Survives (2026d, Appendix J) \\
Contingently tacit & Possession versus transmission: Newton 91st, Leibniz 2nd on the Giant Score & A new composite of unbinding, similarity and path counts & Survives (2026d, Appendix J) \\
Regime-bound tacit & The preference for direct over circuitous lineages & No query returns it; a closed-form law describes it & Dissolves into lexicographic order \\
Outside the substrate & That the mainstream of mathematics ought to rank high & Judgement by historians and mathematicians & Not affected \\
\bottomrule
\end{tabularx}
\end{table}

\subsection{The trade-off}\label{sec:tradeoff}

The preference of the traversal for direct over circuitous lineages is the clearest example of regime-bound tacit information. In Don't Care mode it decides which candidates fill the frontier. Under \CRtwo{} ordering Leibniz ranks second and Euler, J.~Bernoulli, Lagrange and Poisson enter the top 20; under the dictionary's lexicographic ordering their places go to anatomists, surgeons, physicians and humanists, and Leibniz falls to twentieth (\href{https://arxiv.org/html/2605.31470}{2026d}). No query returns ``prefer direct routes'' as an item. And the regime that would make every step exactly inspectable---full rescue---pins \CRone{} at 1.0 and dissolves the preference (2026a; 2026d).

The companion paper on Minsky's Emotion Machine states the general form: audit and valuation are one circuit at opposite settings of one parameter (\href{https://arxiv.org/abs/2607.16623}{2026f}). Repair every collision and the trace is exactly auditable but no path outranks another; tolerate collisions and the \CRtwo{} decay that credits direct routes makes the read approximate. We name this \emph{audit--valuation complementarity} and propose it as the formal counterpart of Polanyi's claim that tacit and explicit knowledge are not sharply divided: one substrate, one dial, two readouts that cannot both be at full strength. The core of the trade-off can be stated exactly.

\begin{lemma}[No valuation without imperfection]\label{lem:novaluation}
Because $\mathrm{CR}_2(n)=\mathrm{CR}_2(n-1)\cdot\mathrm{CR}_1(n-1)$, with $\mathrm{CR}_2$ starting at 1, a regime that repairs every collision, so that $\mathrm{CR}_1=1$ at every step, assigns $\mathrm{CR}_2=1$ to every path, and $\mathrm{CR}_2$ cannot rank candidates. Conversely, two candidates can receive different $\mathrm{CR}_2$ only if $\mathrm{CR}_1<1$ at some step of at least one of their paths. In the traversal of 2026d, where queries are exact stored vectors, $\mathrm{CR}_1<1$ arises from collided blocks abstaining, that is, from tolerated collisions.
\end{lemma}

\begin{conjecture}[Monotone complementarity]\label{conj:monotone}
At fixed memory depth, the dispersion of $\mathrm{CR}_2$ among candidates inside the frontier is non-increasing in the rescue rate RR, and it is zero at $\mathrm{RR}=1$ by Lemma~\ref{lem:novaluation}.
\end{conjecture}

The lemma is elementary, and that is its point: the valuation exists only because of the imperfection that a full audit removes. The conjecture is what \S\ref{sec:predictions} proposes to measure. One qualification keeps the claim honest. The coarse historical structures, such as the Leibnizian hourglass and the Monastery Wall, survive in both modes almost bit-identically; the modes differ in the ranking of candidates inside the frontier, which is exactly where \CRtwo{} adds semantic content (2026d, Appendix J). Explicit structure is robust to the dial. Tacit valuation is fragile to explicitation.

\begin{figure}[tbp]
\centering
\includegraphics[width=\linewidth]{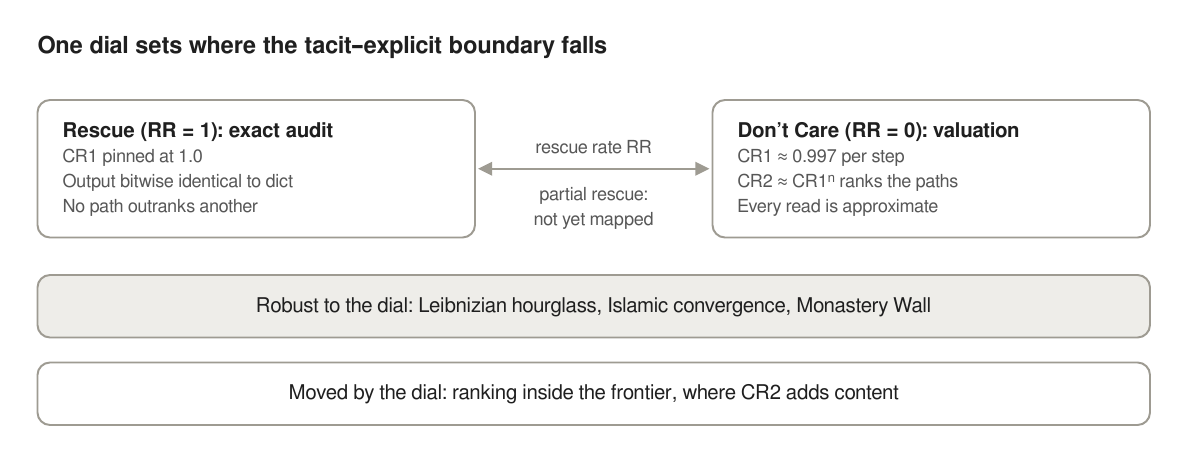}
\caption{One dial, two readouts. Full rescue yields an exact trace with no path valuation; Don't Care yields valuation with approximate reads. The coarse historical structures do not move with the dial; the ranking inside the frontier does. Intermediate rescue rates have not been mapped.}\label{fig:dial}
\end{figure}

\subsection{The pianist, exactly}\label{sec:pianist}

Polanyi observed that a pianist who shifts attention from the music to his fingers may falter (Polanyi 1958). Full rescue is the algebraic analogue. The attempt to make every step of the traversal exactly inspectable removes the small per-step imperfection, $\mathrm{CR}_1\approx 0.997$, whose accumulation was the traversal's skill. The change is not metaphorical: the within-frontier ordering switches from a \CRtwo{} ranking with historical content to a lexicographic one without it. In both cases the skill was never stored as an item; it is a property of how the integration runs, and it returns when the integration is allowed to run again---for the pianist when attention goes back to the music, for VaCoAl/PyVaCoAl when RR goes back to 0.

\subsection{Articulation at the level of law}\label{sec:law}

The valuation is regime-bound, but it is not ineffable in absolute terms. Its magnitude is predicted in closed form: with $\mathrm{CR}_1\approx 0.997$ at sufficient memory depth, \CRtwo{} at generation 56 should be about $0.997^{56}\approx 0.846$, against a measured value of about 0.905 across 25.5 million paths, with the residual consistent with survivor bias along deep paths (2026d). The system cannot say why it prefers direct routes. Its designers can, as a law about the whole regime rather than as an item retrieved from it. This is the structure of Collins's somatic tacit knowledge: the cyclist cannot state the balancing rule, while physics can (Collins 2010). ``We cannot tell'' therefore splits into two statements that are both true at once---the system cannot tell at the object level, and an outside theory can tell at the level of law, where its claim can also fail (\S\ref{sec:falsification}).

Why does the regime produce a valuation at all? Because the search is bounded. On a DAG whose paths grow multiplicatively with depth, an absolute exact match over all paths cannot exist, so VaCoAl/PyVaCoAl builds in a cognitive bound, the frontier size, and ranks the candidates inside it by \CRtwo{} (2026d, Appendix C). This places tacit valuation where Simon placed satisficing: it is what a bounded searcher uses to decide what to keep when it cannot keep everything (Simon 1955).

An objection presses here. If \CRtwo{} is close to a closed-form decay, why not keep full rescue and attach to each path an explicit weight $\gamma^{n}$? The trace would stay exact and direct routes would still be preferred. Such a weight would be arbitrary in a way that \CRtwo{} is not. Its form and its parameter are chosen from outside the memory, and $\gamma$ can only be read off the collision statistics of a Don't Care run, or of a model of them: the stated rule borrows its one number from the regime it is meant to replace. As a ranking, moreover, $\gamma^{n}$ is nothing but hop count, since every $\gamma$ between 0 and 1 orders paths in the same way, and it leaves paths of equal length tied. \CRtwo{} is not chosen. Each factor \CRone{} is the share of blocks that agreed on that retrieval, and the product composes these agreements along the path as independent reliabilities compose along a chain; 2026b reads the same product as multiplicative replay decay in the hippocampus. The two also differ for the third layer of \S\ref{sec:reroot}: $\gamma^{n}$ tells a reader only how long a route is, whereas \CRtwo{} can be taken apart into the \CRone{} of each retrieval along it, so a reader can see where the memory's agreement weakened, which is where judgement is needed (\S\ref{sec:telling}). Nor would an attached weight touch Lemma~\ref{lem:novaluation}, which concerns the valuation that the substrate itself produces: under full audit any valuation that remains must be stated, as $\gamma^{n}$ is. What is grounded in this way is \CRtwo{} as the memory's own confidence in a route. That it also tracks which lineages matter historically is not derived from the algebra; it was found in 2026d and judged by historians (\S\ref{sec:fourpositions}). Whether \CRtwo{} carries anything that hop count does not is already answered across configurations. A traversal ranked by hop count is fixed by the graph and cannot change with the block count $N$, and full rescue makes the output identical to a dictionary whatever $N$ is; yet on the same graph at the same total capacity, $N=128$ settles on a Leibniz-dominant lineage and $N=256$ on a Gauss-dominant one (\S\ref{sec:fourpositions}; 2026d, Appendix F). The dependence on $N$ can only come from tolerated collisions, so in at least one of the two configurations the traversal guided by \CRtwo{} departs from one guided by hop count. Where it departs, at which frontiers, and whether the departures carry historical content has not been measured, and prediction~\ref{pred:sort} states the comparison. Where \CRtwo{} agrees with hop count, the valuation is statable as a preference for shorter routes; the claim of a regime-bound valuation in the substrate rests on the places where it does not.

The point is general, and the rest of the paper leans on it. A large DAG is a reservoir of tacit knowing: the statements that could be drawn out of it have no practical limit, and they cannot be drawn out one by one. To know such a graph is therefore to cull it. No searcher reads every path of a graph whose paths multiply with depth; each keeps a few, and what it comes to know is shaped by whatever does the dropping. In the substrate the ranking that culls is made of tolerated collisions, so there is no informed selection without imperfection: repair everything and the bound is enforced by lexicographic order, which knows nothing of the graph. When the graph exceeds the frontier, what is drawn out first depends on the regime of the cull, and two configurations of one memory can settle on two different lineages (\S\ref{sec:fourpositions}). What is drawn out need not be less exact for that: the regime decides which facts come forward, not whether they are facts. Searchers differ in where their cull comes from and in whether it can be walked back afterwards; \S\ref{sec:culls} compares three.

\subsection{Four positions, one substrate}\label{sec:fourpositions}

The relational definition of \S\ref{sec:def} also explains why the debate of \S\ref{sec:positions} has lasted so long. Each position describes one class of tacitness correctly, and each is narrow in a characteristic way (Table~\ref{tab:four}).

\begin{table}[tbp]
\centering\footnotesize
\caption{The four positions of \S\ref{sec:positions}, read as descriptions of one substrate. Each is right about the class it describes and narrow in a characteristic way.}\label{tab:four}
\smallskip
\begin{tabularx}{\linewidth}{P{0.14\linewidth}P{0.19\linewidth}LL}
\toprule
Position & Class it describes (\S\ref{sec:def}) & What it gets right, in VaCoAl/PyVaCoAl terms & What it leaves out \\
\midrule
Polanyi & Every class at once & Tacit and explicit are not separate stores but two readouts of one memory, set by one dial & What sets the boundary in a given substrate, and what moves it \\
Nonaka \& Takeuchi; Nonaka \& von Krogh & Contingently tacit; since 2009 also an embodied end that cannot be represented & Contingently tacit structure can be externalized by enlarging the family of operations; metaphor, analogy, hypothesis and model act as new queries & Why something stays unstated: one continuum of content cannot tell structure no query has yet reached from structure the stating regime destroys \\
Tsoukas & Regime-bound tacit & Regime-bound valuation cannot be converted at the object level; converting it destroys it & Contingently tacit structure, which a new operation does externalize (\S\ref{sec:externalization}) \\
Collins & Relational: contingently tacit; somatic: regime-bound; collective: outside the substrate & ``Tacit'' names several things, which sit at different strata of the substrate & The dial that trades the somatic stratum against exact audit (\S\ref{sec:tradeoff}) \\
\bottomrule
\end{tabularx}
\end{table}

Nonaka and Takeuchi's conversion is correct for contingently tacit structure. Their externalization works through metaphor, analogy, hypothesis and model, and each of these acts as a new query put to the same memory run the same way (\S\ref{sec:externalization}). Nonaka and von Krogh reject the reading of the theory as treating all tacit knowledge as convertible content: at one end of a continuum lies embodied knowledge that cannot be represented (Nonaka \& von Krogh 2009, 642). A continuum of this kind, which orders knowledge by how far it can be articulated, cannot say why something stays unstated. It may be that no query has yet reached it, which is contingent tacitness, or that every regime that could state it is one in which it no longer exists, which is regime-bound tacitness. Only in the second case does explication cost what it explicates. Tsoukas is right about regime-bound tacitness and narrow in the opposite direction: he denies conversion even for contingently tacit structure.

Collins's account divides the classes most finely, because his three kinds sit at three strata of the substrate, and only two of them lie inside it. \emph{Relational} tacit knowledge is bound structure, tacit for a contingent reason of scale. The Leibnizian hourglass was present in Wikidata before anyone measured it, and standard tools failed on its 25.5 million paths until an operation that could survey the whole graph semantically turned it into stated facts: 47 of 64 lineages through Leibniz, a 10.1-fold thickness ratio (\href{https://arxiv.org/html/2605.31470}{2026d}). \emph{Somatic} tacit knowledge is regime-level valuation, tacit because of what the memory physically is---finite depth per block, residual collisions, the multiplicative structure of \CRtwo---and, like a cyclist's balance, explained from outside in closed form (\S\ref{sec:law}). The ``body'' matters for what the memory comes to prefer: under a fixed total capacity $N\cdot 2^m=2^{34}$, a configuration with $N=128$ settles on a Leibniz-dominant lineage and $N=256$ on a Gauss-dominant one, each historically defensible (2026d, Appendix F). \emph{Collective} tacit knowledge lies outside the substrate. The architecture does not know which lineage matters, because evaluative self-assessment requires value binding that it does not have (\href{https://arxiv.org/abs/2607.16623}{2026f}). Even our claim that the \CRtwo{} ranking is ``historically valid'' is a judgement drawn from the collective tacit knowledge of historians and mathematicians; \S\ref{sec:reroot} makes the community one of the system's layers. What Collins's account does not show is the dial that trades the somatic stratum against exact audit. Polanyi's dictum covers every class at once; what it leaves open is what sets the boundary in a given substrate and what moves it, and that is what the dial of \S\ref{sec:tradeoff} supplies.

\subsection{Simon: all tacitness contingent}\label{sec:simon}

Simon's reply to Polanyi (\S\ref{sec:meno}) carries a position on tacit knowledge, and it can be placed in the same way. For Simon every skill was symbol processing, and intuition was ``nothing more and nothing less than recognition'' (quoted in Kahneman 2011): an expert recognizes a familiar pattern and retrieves what is attached to it. In the terms of \S\ref{sec:def} this admits structural tacitness and treats all of it as contingent. What the expert cannot tell is an item that a better query would state, and what guides a search is a heuristic that a record of the program would state, as a record of Deep Blue's rules would have. On this point Simon and the conversion reading of Nonaka meet from opposite directions: both assume that whatever is tacit is waiting for the right question. What Simon's position leaves out is the second kind of structural tacitness, a valuation that guides a bounded search and exists only while the search runs in the regime that produces it. The substrate shows such a valuation exactly, and the strongest systems built on Simon's reply rely on candidates for it (\S\ref{sec:culls}). Simon asked of the guide of search where its information came from. The question his own bound raises is whether the guide can be stated without being destroyed.

\subsection{A second dualism: System 1 and System 2}\label{sec:twodualisms}

\begin{figure}[tbp]
\centering
\includegraphics[width=\linewidth]{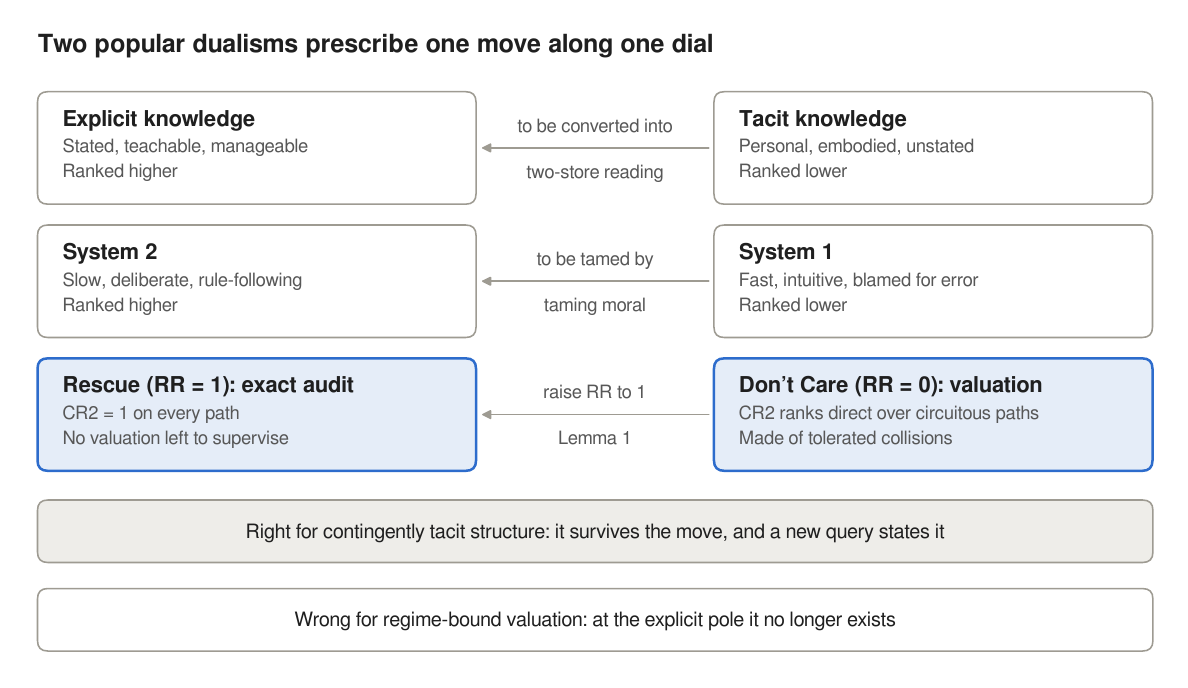}
\caption{Two dualisms, one dial. The two-store reading of knowledge and the taming moral of dual-process psychology prescribe the same move, toward the explicit, deliberate pole. In VaCoAl/PyVaCoAl that move is raising the rescue rate to 1, where Lemma~\ref{lem:novaluation} leaves no valuation to supervise. The move is right for contingently tacit structure and wrong for regime-bound valuation.}\label{fig:dualisms}
\end{figure}

The two-store reading of \S\ref{sec:positions} has three parts: knowledge is split into two kinds of content, the explicit kind is ranked higher because it can be inspected, taught and managed, and the tacit is to be converted into the explicit. The taming moral of \S\ref{sec:dualprocess} has the same three parts: two systems, the deliberate one ranked higher, and System~2 trained to tame System~1. Table~\ref{tab:dualisms} sets the two side by side, and Figure~\ref{fig:dualisms} shows the single move that both prescribe.

The replies coincide as well, part for part. On the split, Polanyi denies that there are two stores: all knowledge is tacit or rooted in tacit knowing (Polanyi 1969, 144). Mercier and Sperber deny that there are two systems: reasoning is intuitive inference, applied to reasons (2017, ch.~9). On the ranking, both turn the dependence round. For Polanyi a maxim guides only someone who already has the art it belongs to (Polanyi 1958, 31, 50); for Mercier and Sperber the step from a reason to what it supports is itself an intuition (2017, ch.~7\nobreakdash--9). On the prescription, both report that enforcing it damages what it was meant to improve. Polanyi's pianist falters under focal attention (Polanyi 1958, 56); Mercier and Sperber's solitary reasoner piles up reasons for the view he started with (2017, ch.~11). We take the two replies to be one critique of one picture, which mistakes two ways in which one activity can run for two things, and then asks which should rule.

\begin{table}[tbp]
\centering\footnotesize
\caption{Two popular dualisms and their common structure. The last column gives the counterpart of each row in VaCoAl/PyVaCoAl.}\label{tab:dualisms}
\smallskip
\begin{tabularx}{\linewidth}{P{0.13\linewidth}LLL}
\toprule
 & Tacit / explicit: the two-store reading & System~1 / System~2: the taming moral & In VaCoAl/PyVaCoAl \\
\midrule
The split & Two kinds of content, a tacit store and an explicit store & Two kinds of mechanism, intuition and reasoning & One memory and one family of operations; two regimes of one dial \\
The ranking & The explicit store is higher: it can be inspected, taught and managed & System~2 is higher: it follows rules and corrects & Neither readout outranks the other; each lacks what the other gives \\
The prescription & Convert the tacit into the explicit & Train System~2 to tame System~1 & Raise the rescue rate to $\mathrm{RR}=1$ \\
The reply & Polanyi: all knowledge is tacit or rooted in tacit knowing & Mercier and Sperber: reasoning is intuitive inference about reasons & Audit is unbinding directed at the system's own trace: the same algebra, not a second system (2026f) \\
What enforcing it costs & The pianist falters; scrutiny of the particulars destroys the whole & Solitary deliberation entrenches the starting view; verbalizing step by step lowers performance on some tasks & Lemma~\ref{lem:novaluation}: $\mathrm{CR}_2=1$ on every path, and ranking falls back to lexicographic order \\
Where it is right & Contingently tacit structure, which a new query states at no cost & Problems with a stateable rule; evaluating other people's arguments & The coarse structures, which survive full rescue (2026d, Appendix~J) \\
Who disowns the hard form & Nonaka \& von Krogh (2009) & Evans \& Stanovich (2013); Kahneman's own caveats (2011) & --- \\
\bottomrule
\end{tabularx}
\end{table}

The substrate states the common point exactly. First, there is no second system: audit is unbinding, the exact inverse of the binding that wrote the trace, run by the same circuit at another setting (2026f). Second, to supervise every step completely is to set $\mathrm{RR}=1$, and Lemma~\ref{lem:novaluation} then gives $\mathrm{CR}_2=1$ on every path: in at least one memory, complete explicit supervision does not improve the valuation it supervises but removes it. Third, the dial does not say ``never audit''. For contingently tacit structure, conversion and deliberate checking do what their advocates say; for regime-bound tacitness they destroy their object. The same sorting appears in people, on Polanyi's own example: witnesses who put a face into words are afterwards worse at picking it out of a line-up (Schooler \& Engstler-Schooler 1990). This is the Identikit of \S\ref{sec:identikit}, measured.

One disanalogy keeps the claim honest. In the substrate the audit readout is exact; in people, on Mercier and Sperber's account, the reflective readout is an advocate rather than an auditor, so the human case is worse for the taming moral, not better. This is why \S\ref{sec:reroot} places audit in an exact layer and valuation in a community that argues.

\subsection{Three ways to cull a DAG}\label{sec:culls}

\begin{figure}[tbp]
\centering
\includegraphics[width=\linewidth]{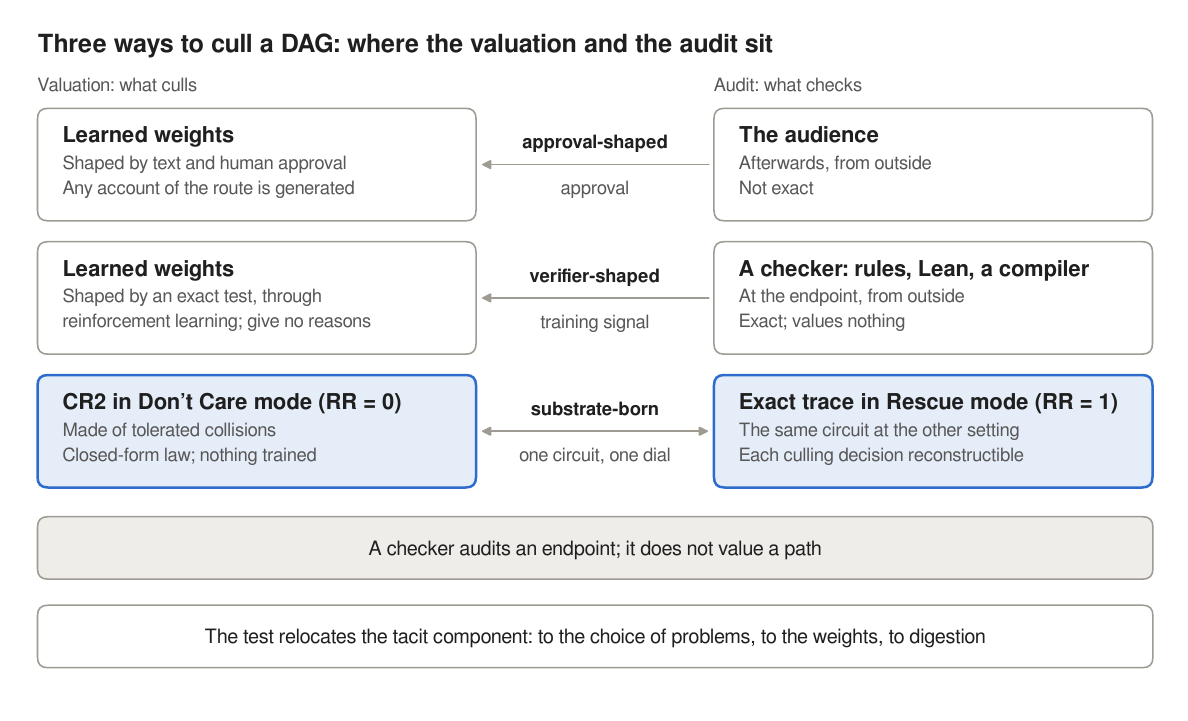}
\caption{Three ways to cull a DAG. In the approval-shaped and verifier-shaped culls the valuation sits in learned weights and the audit stands outside, joined to it only by approval or by a training signal. In VaCoAl/PyVaCoAl the two are one circuit at opposite settings of one dial, so a culling decision can be set beside its exact trace.}\label{fig:culls}
\end{figure}

To know a large DAG is to cull it (\S\ref{sec:law}). Three ways of culling are now in use. They differ in where the valuation comes from, where the audit is done and how the two are connected (Table~\ref{tab:culls}, Figure~\ref{fig:culls}).

\emph{The approval-shaped cull.} A language model trained on text and tuned by human preference culls by a learned sense of what a reader will accept. Its valuation sits in its weights and was shaped by approval. No exact audit takes place inside it, and the account it gives of its own route is one more output (\S\ref{sec:llm}). Whatever checking happens is done afterwards, by the audience.

\emph{The verifier-shaped cull.} In the verifier lineage the valuation again sits in learned weights, but it was shaped by an exact test and not by an audience. The test stands outside the generator and speaks at the endpoint. A proof checker is an object of the Rescue pole. It audits exactly and values nothing: it cannot say that one failed attempt came closer than another. A binary verdict gives a search nothing to climb, so a graded signal has to be made out of something else. AlphaGo made it with a value network (Silver et al.\ 2016). AlphaProof Nexus makes it with rater models that compare proof sketches in pairs and aggregate their verdicts into Elo ratings that steer the search (Tsoukalas et al.\ 2026). The audit is exact and external, the valuation learned and internal.

\emph{The substrate-born cull.} In VaCoAl/PyVaCoAl the ranking that culls is \CRtwo, a by-product of the collisions the memory tolerates as it traverses (\S\ref{sec:tradeoff}). No audience shaped it and no test trained it. It is produced by the circuit that audits exactly at the other setting of the dial, so every culling decision can be reconstructed afterwards and set beside the exact read (2026g).

\begin{table}[tbp]
\centering\footnotesize
\caption{Three ways to cull a DAG. The rows ask where the valuation comes from, where the audit is done, whether a culling decision can be walked back, and where the tacit component sits.}\label{tab:culls}
\smallskip
\begin{tabularx}{\linewidth}{P{0.15\linewidth}LLL}
\toprule
 & Approval-shaped & Verifier-shaped & Substrate-born \\
\midrule
Example & A language model tuned by human preference & AlphaGo, AlphaGeometry, AlphaProof, AlphaProof Nexus & VaCoAl/PyVaCoAl in Don't Care mode \\
What shaped the valuation & Human text and human approval & An exact test, through reinforcement learning & Tolerated collisions during traversal; nothing was trained \\
Where the valuation is kept & Learned weights & Learned weights: policy and value networks, rater models & \CRtwo{} on each path, with a closed-form law (\S\ref{sec:law}) \\
Where the audit is done & Outside and afterwards, by the audience; not exact & Outside and at the endpoint, by a checker; exact & Inside, by the same circuit at $\mathrm{RR}=1$; exact \\
Can a culling decision be walked back? & No: the reported route may differ from the one taken (\S\ref{sec:llm}) & The search record can be walked; the judgements on it come from networks that give no reasons & Yes: each ranking decision can be reconstructed (2026g) \\
Where the tacit component sits & In the weights and in the audience & In the choice of problems, in the weights and in the digestion still owed & In the regime, as regime-bound valuation, and in the community that values \\
Typical product & Persuasive and unverified & Verified and rootless (\S\ref{sec:verifier}) & Traced, with a confidence; not yet valued \\
\bottomrule
\end{tabularx}
\end{table}

The comparison is not a ranking: the verifier-shaped cull has found proofs that people had sought for decades, and nothing built on the substrate has. The first cull has valuation without exact audit; the second has both and keeps them apart; the third has both in one circuit and pays for it, since the two cannot be at full strength together (Lemma~\ref{lem:novaluation}).

The dispute of \S\ref{sec:meno} can now be given a verdict in which each side is right about a part. Simon is right that a test and a generator suffice to solve problems. Polanyi is right that the test does not remove the tacit component. It moves it, to three places. It moves it upstream, into the choice of which problems are given a formal statement: the Nexus results came from open problems that people had selected and formalized, and experts had to check that the formal statements said what the conjectures say (Tsoukalas et al.\ 2026). It moves it inside, into the weights that hold the generator's sense of where to look: the machine counterpart of Polanyi's intimation. And it moves it downstream, into the digestion that a certified result still needs before anyone can use it (\S\ref{sec:verifier}). A checker settles justification; the cull belongs to discovery, and to a further question that no test addresses: what a result is for.

A defender of Simon will reply that the weights are numbers, every one of which can be printed. Nothing about them is ineffable, but a printed weight is checkable, as a formal proof is, without being walkable in the sense of \S\ref{sec:rootless-def}: no one can find in the printed weights the judgement that dropped a given branch, change it for a reason and see what the search would then have kept. Simon's heuristics were meant to be statable in this stronger sense, as rules a person could apply. That is the half of his reply that the strongest systems do not bear out.

\subsection{Externalization with a trace}\label{sec:externalization}

The programme has already externalized contingently tacit structure several times by adding operations. The ``calculus degree'' and the Giant Score are new compositions of binding, bundling, unbinding and Hamming similarity (2026a; 2026d). They turned a feature of the graph into a stated fact: Newton and Leibniz have nearly equal calculus-content scores (0.534 and 0.569) but very different path counts to the 64 Fields Medalists (4 and 47), so Newton ranks 91st on the Giant Score and Leibniz second, the difference between possessing and transmitting upper-tail human capital (2026d). Nonaka and Takeuchi located externalization in metaphor, analogy, hypothesis and model; hyperdimensional operations are a computational form of those moves. What VaCoAl adds is the trace. Because binding is exactly reversible, every externalized item arrives with its derivation---which bundles, which roles, which paths---so a result that looks like magic can be taken apart afterwards by anyone who holds the substrate.

\subsection{Two axes: how a store is written and how it is read}\label{sec:stores}

The dial of \S\ref{sec:tradeoff} describes how a store is read. How a store is written is independent of it; in 2026c we put the point in one line: ``what cannot be exactly decomposed was never composed, only mixed.'' A store is \emph{composed} when its items are bound records that keep who stands in which role and can be taken apart exactly, as in an ontology or an edge list; it is \emph{mixed} when its items are superposed and found again by similarity, as in a vector index. ``The dog bites the man'' and ``the man bites the dog'' are different records under directed binding (\S\ref{sec:fromto}) and one point in a store that keeps only the sum of its parts. Crossing the two axes gives Table~\ref{tab:stores}: VaCoAl/PyVaCoAl holds one composed store and reads it either way, while a proof checker reads a composed object exactly and in no other way.

\begin{table}[tbp]
\centering\footnotesize
\caption{Two axes. The columns give how a store is read, the rows how it is written. The placing of language-model weights is a hypothesis of this paper.}\label{tab:stores}
\smallskip
\begin{tabularx}{\linewidth}{P{0.20\linewidth}LL}
\toprule
 & Read exactly & Read with tolerance \\
\midrule
Composed store: bound records, as in an ontology or an edge list & Rescue mode: an exact trace and no valuation. A formal proof under a proof checker. & Don't Care mode: a \CRtwo{} valuation that can be set beside the exact trace of the same paths. \\
Mixed store: superposition, as in a vector index & Not available: there is nothing to decompose, so any account of a route is generated like the output itself. & Similarity completion: fluent and graded, and not walkable. The weights of a language model, largely. \\
\bottomrule
\end{tabularx}
\end{table}

The weights of a language model are, we suggest, largely a mixed store. Models trained on ``A is B'' fail to retrieve ``B is A'' (Berglund et al.\ 2024). Facts learned from separate documents do not compose inside the weights (Balesni, Korbak \& Evans 2024). And small networks hold more features than they have dimensions by superposing them, at the price of interference (\href{https://arxiv.org/abs/2209.10652}{Elhage et al.\ 2022}). The claim concerns what is stored and found again, not what is processed in context: the first two failures disappear when the facts are in the prompt.

Two hypotheses follow. The first concerns tuning. An approval-shaped cull and a verifier-shaped cull are two ways of shaping the readout of a learned store; neither changes the way the store is written. Models of the two kinds should therefore differ where their anchors---a checker, an audience---are present, and fail alike where the anchors are absent (\S\ref{sec:verifier}; prediction~\ref{pred:tuning}). The second is a guard against a misreading. The two kinds of store are not the two stores of \S\ref{sec:twodualisms}. Tacit is not ``vector'' and explicit is not ``ontology'': the regime-bound valuation of \S\ref{sec:tradeoff} arises in a wholly composed store, read with tolerance. The way a store is written decides whether what is drawn out of it can be walked back: the tacit component of a composed store is tacit by regime; that of a mixed store is tacit by construction. Whether one system can have both the fluency of a mixed store and the exact trace of a composed one we leave open (\S\ref{sec:open}); the hippocampus, as 2026b describes it, keeps two pathways under gain control.

\section{The mentor--student DAG as a tacit-knowledge transmission graph}\label{sec:dag}

The graph our programme traversed is not a graph of explicit knowledge. Its edges are apprenticeships, the channel Polanyi named for passing on what cannot be specified, so its macro-structures are events in the history of tacit-knowledge transmission.

\subsection{What an edge carries}\label{sec:edge}

Each edge asserts that one person trained another; the relation merges Wikidata's doctoral-advisor, student and student-of properties (\href{https://arxiv.org/html/2605.31470}{2026d}, Appendix A). Polanyi held that an art which cannot be specified in detail is transmitted by example from master to apprentice (Polanyi 1958). An apprenticeship edge is therefore a socialization channel in Nonaka's sense, and the topology of the graph records the routes by which tacit knowledge travelled. The nodes carry something else. Field, language, employer, society membership and dates are explicit attributes, bound into each hub's composite vector (2026d, Appendix C). The two layers can be read separately: the topology cannot see the medieval translation movement, which moved knowledge through books rather than masters, but unbinding the FIELD component of the node vectors recovers it as an astronomy signal (2026d). A DAG memory of this kind thus holds both channels at once: tacit transmission in its edges, codified content in its node vectors.

\subsection{The hourglass as institutionalized socialization}\label{sec:hourglass}

Leibniz built an apparatus of transmission: he founded the Berlin Academy of Sciences in 1700, kept one of Europe's largest scholarly correspondences, and started a lineage running through Johann Bernoulli, Euler, Lagrange, Laplace and Poisson (2026d). In the graph this appears as an hourglass, 5.3 paths per node upstream against 53.4 downstream (2026d). Across the same window learned-society membership rose from 0.08 to 3.81 memberships per person (\href{https://www.alphaxiv.org/abs/2608.17915}{2026g}; 2026d). Academies, correspondence and lineages are institutions for multiplying socialization. They are, nearly item for item, the follow-up activities Tao lists---talks, meetings with experts, workshops, newcomers joining a field.

\subsection{Newton, Thurston, and the follow-up gap}\label{sec:newton}

Newton and Leibniz have nearly equal calculus-content scores in the graph, but only four of the 64 lineages pass through Newton, against 47 through Leibniz; Newton worked largely in isolation and trained few successors (2026d). Newton possessed the calculus; Leibniz possessed and transmitted it. The graph records only documented mentorship, not Newton's abundant influence through texts, but mentorship is exactly the channel of tacit knowledge. Thurston recorded a similar precedent (\href{https://arxiv.org/abs/math/9404236}{Thurston 1994}). After his classification theorem for foliations, the field began to empty within a couple of years: his papers set down only the key points, many students and mathematicians struggled to understand the proofs and left, and he concluded that ``more than the knowledge, people want personal understanding.'' The follow-up gap is therefore not new. What is new is the rate at which answers can be produced without anyone acquiring the understanding that used to come with them. The AI-solved results Tao describes are the limiting case: possession without any human possessor, and therefore without the apprenticeship edges along which understanding normally flows.

\subsection{The Monastery Wall as a change of medium}\label{sec:wall}

Every traversal stops at the 57th generation, at Lanfranc of Canterbury (1005--1089), who taught at Bec, or at Ivo of Chartres (1040--1116), whose own teachers are unrecorded (2026d). Before them, education was organized in monastic and cathedral schools as a communal affair, and such channels produced no structured mentor--student records (2026d; 2026g). In Collins's terms the Wall is a shift in the medium of tacit transmission, from a communal medium close to collective tacit knowledge to a dyadic medium of apprenticeship. It is not an absence of transmission. It is the point at which transmission began to leave the kind of trace our instrument can read.

\subsection{Is a second wall forming?}\label{sec:aiwall}

If the follow-up gap widens, a future prosopography of twenty-first-century science could meet a boundary of a new kind. Results would be attached to nodes that have few or no human apprentice edges, because what was learned in producing them was learned by no person. We call this hypothetical boundary an \emph{AI wall}. Unlike the Monastery Wall, it would mark not a change of human medium but a channel in which no human holds the tacit root. Prediction~\ref{pred:wall} gives a measurable signature. This is a prediction, not an observation, and the re-rooting of \S\ref{sec:reroot} is what would prevent it.

\subsection{The Kuhnian shift as a change of regime}\label{sec:kuhnshift}

Kuhn located the deepest sense of a paradigm in exemplars, concrete solutions through which students learn to extend a field, and he tied that learning to tacit knowledge (Kuhn 1970, Postscript, \S\S3--4). Exemplars travel along apprenticeship edges, so a change of paradigm should leave a trace in the graph: the descendants of the old exemplars thin out, and lineages that did not descend from them enter. The trace is not the shift. In the terms of \S\ref{sec:def}, a shift changes what a community asks and how it reads what it has: it is a move from one regime and family of queries to another, and to a new vocabulary in which records are written (\S\ref{sec:schema}). A re-ranking of lineages under a fixed regime is normal science; a change of regime is what Kuhn called a revolution.

The distinction matters because \CRtwo{} is, by construction, an instrument of normal science. Because every step multiplies a path's confidence by a factor below one, the traversal credits direct routes, which makes it good at tracing a tradition and unable to see a revolution coming: a lineage that does not descend from the current exemplars reaches the frontier, if at all, by long paths, and it is ranked low for the very reason that makes it new. We do not propose to repair this by changing the valuation, which would lose what it is good for. Two things can be set beside it. The first is an \emph{anomaly ledger}. A result that the community does not accept, or that changes when the memory is configured differently (\S\ref{sec:fourpositions}), is recorded with its query, regime and date instead of being discarded, so that the accumulation of anomalies Kuhn described (Kuhn 1970, chs.~6--8) becomes a public record. The second is the dial. By Conjecture~\ref{conj:monotone}, a little more rescue flattens the preference for direct descent, so the rescue rate may act as a control between the exploitation of what is known and the exploration of what is not. For the dial to explore, the ties it creates must be broken at random, not by lexicographic order, and any rise in the diversity of lineages must exceed what random sampling from the same frontier gives; and even then it varies without selecting, since it cannot tell a promising heresy from a merely unusual route. Kuhn held that sustaining the tension between convergent and divergent research is a prime requisite of the best research (Kuhn 1977, ch.~9); in the substrate the balance would be a setting that can be declared.

The question is not idle, because the collective record already leans toward exploitation. In 41.3 million papers across the natural sciences, scientists who used AI tools published 3.02 times as many papers and received 4.84 times as many citations as those who did not, while the collective volume of topics studied shrank by 4.63 percent and scientists' engagement with one another fell by 22 percent (Hao et al.\ 2026). Deployed without a ledger, with its dial left at one pole, the instrument we propose could add to the tilt it is meant to make visible. We state this as a risk of our own design.

\section{Rootless explicit knowledge: a Polanyian diagnosis of the follow-up gap}\label{sec:rootless}

An AI result that is verified but held tacitly by no one is explicit knowledge without a root, and the follow-up gap is what rootlessness looks like from outside.

\subsection{Definition}\label{sec:rootless-def}

Board games make the definition easy to picture. In March 2016, in its second game against Lee Sedol, DeepMind's AlphaGo played Move~37 on the fifth line, and went on to win the match 4--1 (\href{https://www.technologyreview.com/2026/10/02/1145639/dont-be-fooled-llms-dont-reason/}{Graepel 2026}). We use this move, and IBM's Deep Blue at chess before it, to illustrate three conditions. When all three hold together, we call a result $R$ \emph{rootless}.
\begin{enumerate}
\item \textbf{$R$ is explicit: it is stated openly, and others can check it.} Move~37 was a stone in full view, tested by the rest of the game.
\item \textbf{There is no record of the route to $R$ that a person can follow step by step.} Here the machines differ. Deep Blue, which beat Garry Kasparov in 1997, took far too many steps for anyone to read, but each applied rules written in human terms. AlphaGo's networks supplied hunches about which moves looked promising and which positions looked won, and its search kept a record, a game tree in which every move carried its networks' judgments; the tree can be walked, but the judgments on it come from networks that give no reasons (Graepel 2026). A large language model typically keeps no such record, and the chain of thought it writes out is often constructed after the fact (\S\ref{sec:llm}). Condition~2 is therefore a matter of degree.
\item \textbf{No one yet has the tacit understanding of $R$ (the feel for why it works) needed to extend it, teach it, or judge what it is for.} When Move~37 was played, it had almost no roots in human play: AlphaGo's own policy network, trained on the moves of strong human players, put the chance that an expert would play it at roughly one in 10,000 (Graepel 2026).
\end{enumerate}

Polanyi's dictum implies that a rootless result is not yet complete knowledge. If explicit knowledge must be tacitly understood and applied, a result that nobody understands in that way is true but idle: it can be cited, but not used as knowledge is used. Condition~2 is the one a substrate can change; conditions~1 and~3 belong to the result and to the community respectively.

Condition~2 turns on what it is to follow a route, and we use the word \emph{walkable} for it throughout. A trace is walkable when a person can travel it by doing it: re-enact each step, change one step and see what follows, and carry the route to a problem it was not made for. A route can be checked line by line without being walkable, as a formal proof shows (\S\ref{sec:verifier}). It is the condition under which Polanyi's indwelling (\S\ref{sec:indwelling}) and Nonaka and Takeuchi's internalization take place. It also asks for more than the route that survived. To change a step and see what follows, one has to know what the step was chosen over: which branches were dropped, and by what valuation. A walkable trace therefore includes a record of the cull (\S\ref{sec:culls}). A final proof gives the survivor alone. VaCoAl/PyVaCoAl keeps the paths inside its frontier with their \CRtwo{} ranks beside the exact trace, and the same graph can be read again under another configuration to show what a different body would have kept (\S\ref{sec:fourpositions}). Such a record makes practice possible. It does not do the practice, which remains with people (\S\ref{sec:byhand}).

Pritchard draws a line between knowing and understanding: a student can know Pythagoras' theorem on a teacher's testimony while lacking any grasp of why it holds, and it is understanding that education aims at (Pritchard 2023, 195\nobreakdash--197). Ten years after Move~37, Go shows this line at work, with the AI in the teacher's place. Professional players now train by copying AI moves as closely as they can, ``even when the machine's thinking remains mysterious to them'' (\href{https://www.technologyreview.com/2026/02/27/1133624/ai-is-rewiring-how-the-worlds-best-go-players-think/}{Kim 2026}). A rootless result is known in the first way: it can be checked or accepted, but no one yet understands it well enough to extend, teach or judge it.

\subsection{A worked case}\label{sec:case}

The Jacobian counterexample satisfies all three conditions at the moment of its announcement. Tao noted that it can be verified by a brief calculation, but that the vanishing of every non-constant coefficient of the Jacobian looks like a massive cancellation---1,329 equations against 360 degrees of freedom for a generic degree-seven map---so the example was very unlikely to have been located by brute force (\href{https://terrytao.wordpress.com/2026/07/21/a-digestion-of-the-jacobian-conjecture-counterexample/}{Tao 2026b}). His digestion rebuilds the construction from the multiplication of low-degree polynomials and keeps the remaining ``miracles'' to a minimum, while noting that one of them still lacked a fully satisfactory geometric explanation. The result was verified as explicit knowledge, and its tacit root had to be grown afterwards by human digestion. Sanderson's phrase is the precise statement of condition~2: every AI-generated proof begins life as an unsolved exposition problem (\href{https://terrytao.wordpress.com/2026/09/18/if-math-is-more-than-proof-we-need-to-better-celebrate-the-rest-of-it/}{Sanderson 2026}).

\subsection{Why LLM results are rootless by default}\label{sec:llm}

Four lines of evidence explain why results produced by large language models tend to arrive rootless.
\begin{enumerate}
\item \textbf{The derivation exists, but it is not a walkable trace.} Interpretability tools can recover parts of a transformer's forward pass as a graph: in Claude 3.5 Haiku, the question about the capital of the state containing Dallas activates an intermediate representation of Texas before the answer Austin; replacing it with California changes the answer to Sacramento. The method captures only a fraction of the computation, and understanding the circuits of a prompt of a few dozen words currently takes hours of human effort (\href{https://www.anthropic.com/research/tracing-thoughts-language-model}{Anthropic 2025a}). The DAG is there; the trace is not.
\item \textbf{The implicit DAG is directional and patchy.} Models fail at two-hop questions whose facts were learned from separate documents (\href{https://arxiv.org/abs/2411.16353v1}{Balesni, Korbak \& Evans 2024}), and models trained on ``A is B'' do not generalize to ``B is A'' (\href{https://arxiv.org/abs/2309.12288}{Berglund et al.\ 2024}): relations run one way and compose unreliably inside the weights, in contrast to binding that is exactly invertible (\S\ref{sec:programme}).
\item \textbf{Self-explanation imitates human externalization.} Asked how it added 36 and 59, Claude describes the schoolbook carrying algorithm, while the traced mechanism combines a rough-magnitude path with an exact last-digit path (Anthropic 2025a). When a prompt is biased toward a particular answer, models tend to give that answer and to rationalize it step by step without mentioning the bias (\href{https://arxiv.org/abs/2305.04388}{Turpin et al.\ 2023}). In Polanyian terms, the model's explanation is a borrowed externalization, detached from its own integration.
\item \textbf{Persuasion outruns truth.} Five state-of-the-art assistants trained on human preference consistently told users what they wanted to hear, and both humans and preference models preferred convincingly written sycophantic responses over correct ones a non-negligible fraction of the time (\href{https://arxiv.org/abs/2310.13548}{Sharma et al.\ 2024}). An answer shaped by approval is an argument addressed to an audience, in Mercier and Sperber's sense, rather than a report of how it was reached.
\end{enumerate}

LLMs can produce abductive hypotheses---Peirce's term for the only inference that introduces a new idea (Peirce 1903)---by composing knowledge along implicit paths. But each hypothesis arrives without a walkable trace, with an explanation that may imitate human explanation rather than report the process, and with a persuasiveness tuned to its audience rather than to its support. It is rootless by default. Rootlessness is a default, not a law: a result stops being rootless as soon as someone acquires the understanding needed to extend it, teach it and judge it (\S\ref{sec:byhand}).

\subsection{Verified but rootless: the verifier lineage}\label{sec:verifier}

It might be thought that this default belongs to models tuned by approval, and that a model disciplined by an exact test escapes it. The three conditions say otherwise. A verifier strengthens condition~1 to its limit and leaves conditions~2 and~3 where they were.

AlphaProof Nexus is the clearest case to date (Tsoukalas et al.\ 2026). Its results meet condition~1 as fully as a result can: each proof is accepted by Lean. On condition~2 the record is long and not walkable: the search leaves a population of rated sketches, and the final proof is a formal object that a machine checks line by line. A formal proof is a route that a checker can follow. That does not make it a route that a person can follow: AlphaGeometry's solution to one Olympiad problem runs to more than a hundred deduction steps, many of which its authors call ``extremely tedious to a human reader'' (Trinh et al.\ 2024). Condition~3 is untouched. At the moment the compiler accepts a proof, nobody holds the result tacitly.

The normal product of the verifier-shaped cull is thus a result that is \emph{verified but rootless}: the Jacobian counterexample, made routine. A loop that used to pass through people now closes without them: generator, checker, training signal, generator. While human reading was the only way to confirm a result, understanding came as a by-product of checking; an exact checker confirms without the by-product. In this respect the follow-up gap may be wider behind a verifier than behind an approval-shaped generator: a persuasive, unverified claim invites people to check it, and a certified claim does not. The Nexus authors had experts review formal statements and proofs (Tsoukalas et al.\ 2026): digestion done by hand, which the design does not require.

The lineage also has a boundary, which we call the \emph{verifier's streetlight}. An exact test exists only where the rules are fixed beforehand, as in games, formal mathematics and programs with tests. Most of what scholars argue about has no checker. There a model has nothing to be disciplined by except its audience, and its cull is approval-shaped whatever its training elsewhere may have been.\footnote{In November 2026 the first author gave two kinds of task to general-purpose assistants from two vendors: reading the optical figures of patents and conference papers on lithography, and revising the present paper. One assistant did the first well and the second poorly, and the other the reverse; each was fluent where it was wrong. We do not know how either was trained. We report this as the observation that prompted the hypothesis of \S\ref{sec:stores}, not as evidence for it.} Agentic systems such as POPPER now design and run statistical falsification tests where no checker exists (Huang et al.\ 2025). They widen the streetlight, but which implications are put to the test is chosen by the agents that run the tests, so conditions~2 and~3 stay where they were. Inside the streetlight, results arrive verified and rootless, and the second and third layers of \S\ref{sec:reroot} are needed for understanding. Outside it, results arrive unverified as well, and the same two layers are needed for trust.

\subsection{Rootless forms: art, craft and skilled work}\label{sec:forms}

The three conditions do not mention mathematics. Zoran's analysis of generative AI in art shows the same structure where the result is a form rather than a proof (Zoran 2026). Artistic forms ``stabilize and transmit meaning through artifacts''; artistic becoming is the transformation that creative engagement may bring about in a person or a community, and generative systems produce such forms without that becoming. A form can circulate, but becoming ``cannot circulate at all; it can only be undergone.'' Skilled work shows the same decoupling. The adenoma detection rate of experienced endoscopists in colonoscopy without AI fell from 28.4 to 22.4 percent after AI assistance was introduced into routine practice (Budzy\'{n} et al.\ 2025), and developers who learned a new library with AI assistance scored about 17 percent lower on a test of conceptual understanding, code reading and debugging (Shen \& Tamkin 2026). In both cases the form was delivered, while the becoming that would let the practitioner produce it unaided was thinned. Decoupling, like rootlessness, is a default and not a law: Shen and Tamkin found patterns of AI use under which learning was preserved, and in them the person goes through the route instead of receiving its endpoint, which is what \S\ref{sec:reroot} is designed to make possible. Table~\ref{tab:domains} sets the cases side by side.

\begin{table}[tbp]
\centering\small
\caption{Rootless results across domains: an explicit form arrives while its root is missing.}\label{tab:domains}
\smallskip
\begin{tabularx}{\linewidth}{P{0.13\linewidth}LLP{0.18\linewidth}}
\toprule
Domain & The explicit form & What is missing & Source \\
\midrule
Mathematics & A counterexample to the Jacobian conjecture, verifiable by a brief calculation & A route a person can follow; the understanding needed to extend and teach it & Tao 2026b; Sanderson 2026 \\
Go & Move~37 & The general principles behind AI moves & Graepel 2026; Kim 2026 \\
Art & Forms of a sophistication that once required the maker's becoming & The maker's becoming & Zoran 2026 \\
Clinical skill & Adenomas detected with AI assistance & Unaided detection (28.4\% to 22.4\% without AI) & Budzy\'{n} et al.\ 2025 \\
Programming & Working code in a new library & Conceptual understanding, code reading and debugging (about 17\% lower) & Shen \& Tamkin 2026 \\
\bottomrule
\end{tabularx}
\end{table}

\subsection{Scooping, possession and transmission}\label{sec:scooping}

Tao's last symptom is that promising open directions are withheld for fear of being scooped (Tao 2026a). The historical analysis of \S\ref{sec:dag} suggests why this matters. When generation is cheap, the scarce assets are a direction and the tacit sense of why it is worth pursuing, so people guard them. Guarding them, however, starves transmission, which is what 2026d found to separate Leibniz's legacy from Newton's in the documented record. A credit system that rewards possession pushes a field toward Newton's thin thread; one that rewards transmission pushes it toward Leibniz's hourglass. Sanderson's proposal to credit motivated explanations can be read as an attempt to reward transmission again (Sanderson 2026). Digestion is a public good: walking a route, recording what was dropped and stating which reading was committed to cost the person who does them, while the benefit is spread across everyone who later extends, teaches or corrects the result. The records proposed in this paper are, among other things, objects to which credit can attach: a committed reading (\S\ref{sec:commitment}), a retraction with its reason (\S\ref{sec:falsification}) and a change of schema opened to comment (\S\ref{sec:schema}) can each be published, dated and cited.

\subsection{Falsification and paradigms after the verifier lineage}\label{sec:falsification}

Popper asked whether a statement is open to refutation (Popper 1959); Kuhn asked what a community must share for a result to count, to be extended and to be taught (Kuhn 1970). Both took for granted that new explicit knowledge is drawn out of a record under a regime that a community holds and can follow. This subsection restates the diagnosis in their terms; nothing in \S\ref{sec:reroot} depends on it.

The three conditions of \S\ref{sec:rootless-def} divide between the two vocabularies. Condition~1 is Popper's: a result is stated so that it could fail. Condition~3 is Kuhn's: the competence to see what a result is for, which he located in exemplars learned in practice. Condition~2 lies between them. A checker is falsification made exact and automatic. Popper set the conceiving of a theory outside the logic of science, as a matter for psychology (Popper 1959, \S2), and Simon answered that discovery has a logic of its own (Simon 1973). The verifier lineage draws Popper's line exactly, and adds that the side he set aside is the side on which rootedness is decided. A verified result has survived refutation and can still be idle in Polanyi's sense.

The three places to which \S\ref{sec:culls} found the tacit component moved are, item for item, three things a paradigm used to supply. Upstream, the choice of problems is what normal science took from its paradigm, which tells a community which puzzles are worth solving (Kuhn 1970, ch.~4). In the weights, the generator's sense of where to look is the machine form of the learned similarity by which a student comes to see a new problem as like one already solved (Kuhn 1970, Postscript, \S3). Downstream, digestion is what turns a certified result into an exemplar from which others can learn. Of the values by which a community chooses among theories, fruitfulness, a result's capacity to open further work (Kuhn 1977, 320--339), is what the follow-up gap measures, and it is the value that no checker assesses.

Beyond the endpoint a failed test does not say where it failed. Logic condemns the conjunction of hypothesis and auxiliary assumptions without saying which member to give up, the Duhem--Quine problem; on a DAG a wrong endpoint shows that some edge upstream is wrong, not which. The theory of belief revision fixes the contraction once beliefs are ordered by how firmly they are entrenched, the least entrenched being given up first; beliefs that are equally entrenched are given up together (G\"{a}rdenfors \& Makinson 1988). Three orderings are available in the substrate, and each does its own job. \emph{Provenance} is stated and coarse. Edges that record an observation, a citation or a generator's output are written as different kinds of record, so that a two-step generated shortcut does not outrank a five-step chain of experiments; provenance survives every regime, full rescue included. \emph{\CRtwo{}} is bound to the regime and fine. It separates records of the same stated kind, but what it measures is directness of descent, not weight of evidence, so retraction ordered by it gives up the atypical and circuitous edges first and keeps the core: Lakatos's protective belt, built into the order (Lakatos 1970). It therefore does better than a random tie-break only where errors are more likely on the atypical and circuitous edges than in the core, and worse where the core is wrong, which is the case the ledger is there to detect. The \emph{anomaly ledger} watches the other two. It is the ledger of \S\ref{sec:kuhnshift}, which now records each retraction as well: the failed test, the records withdrawn, and the records that the test also bore on but that the ordering left in place, which are what the retraction protected. Two counts can then be read from it: how often the same record has been protected, and how many of the adjustments made to protect it went on to yield a new prediction that was borne out. A record protected again and again by adjustments that yield nothing new marks a programme degenerating in Lakatos's sense. That is the case in which ordering by \CRtwo{} is wrong, and the ledger lets it be seen as a count rather than only felt. It is the signal to revise the order itself, by a change of regime that the third layer declares (\S\ref{sec:layers}): for instance, to withdraw the protected record on trial and rerun the tests that failed. Whether the revision is warranted remains a judgement; the ledger only says where one is due. Lemma~\ref{lem:novaluation} then says something about falsification. Under full rescue only what has been stated survives, and the coarser the order, the more sound records fall with the faulty one; the finer grades exist only where imperfection is tolerated. We put forward, as an interpretive conjecture and not a result, that falsification beyond the endpoint needs, besides what can be stated, the kind of valuation that Polanyi called tacit. Popper's own account leaves room for this, since the basic statements by which a theory is tested are accepted by a decision, like a jury's verdict (Popper 1959, \S\S29--30). The substrate does not remove the decision; it records which reading was tested, under which regime, and what was given up, marking what is withdrawn rather than deleting it. Each record also keeps its justification, the records from which it was derived, so that a retraction propagates to everything that rested on it, as in a justification-based truth maintenance system (Doyle 1979); in VaCoAl/PyVaCoAl this is cheap, since binding is exactly reversible and the constituents of a composite record can be recovered from it (\S\ref{sec:programme}).

\begin{table}[tbp]

\centering\footnotesize
\caption{What remains open to refutation under the three culls of \S\ref{sec:culls}. ``Refutable'' means that a stated item could turn out wrong by a procedure others can repeat.}\label{tab:falsify}
\smallskip
\begin{tabularx}{\linewidth}{P{0.15\linewidth}LLL}
\toprule
 & Approval-shaped & Verifier-shaped & Substrate-born \\
\midrule
The result & Only by an audience, afterwards; not exact & By an exact checker, for the statement as formalized & By whatever test the domain offers; the substrate adds none \\
The cull & Not refutable: the valuation gives no reasons, and the reported route may differ from the one taken (\S\ref{sec:llm}) & Not refutable: the search record can be walked, the judgements on it cannot & Reconstructible: each ranking decision can be set beside the exact trace (2026g) \\
The reading & A generated explanation, itself one more output & The formal statement, checked by experts against the conjecture (\S\ref{sec:culls}) & A committed reading recorded as bound records; a wrong one is a located error (\S\ref{sec:commitment}) \\
The regime & Changes with tuning, without a record (\S\ref{sec:stores}) & Fixed by the training signal; not stated & Declared ($\rho$); its law predicted in closed form (\S\ref{sec:law}; Conjecture~\ref{conj:monotone}) \\
What orders retraction & Nothing stated; the audience decides afresh each time & Nothing beyond the endpoint; which formal statement to revise is left to people & Provenance, coarsely, as stated; within a kind, \CRtwo{} under tolerated collisions; under full rescue, provenance and then lexicographic order (Lemma~\ref{lem:novaluation}) \\
Where falsification now applies & The audience's scrutiny of the endpoint & The endpoint & The endpoint; the committed reading, which aims a test of the world; the regime's law, which tests the instrument \\
\bottomrule
\end{tabularx}

\end{table}

Table~\ref{tab:falsify} sets out what remains open to refutation under each cull. The cull is not refutable as it stands in either learned case, because the valuation that made it gives no reasons. Two things can be made refutable in its place. One is a committed reading (\S\ref{sec:commitment}), which says which records a failed prediction bears on, so that a test of the world, which still comes from outside, can be aimed. The other is a regime's law (\S\ref{sec:law}). The closed-form decay of \CRtwo{} could have missed the 25.5 million measured paths, and Conjecture~\ref{conj:monotone} names the curve that would refute the dial: what cannot be told at the object level can still be falsified at the level of law. A refuted law, however, refutes the instrument, not a claim about the world. Falsifiability thus moves from the single result to the reading, to the ordering of retraction and to the regime, and it requires the regime to be declared. The substrate adds no test of the world, only a place to aim one and a record of what a failure cost.

The case that neither vocabulary was built for is a regime that changes without notice. By \S\ref{sec:def}, the class of a piece of information depends on the regime $\rho$ and the queries $Q$. Call a move from $(\rho,Q)$ to $(\rho',Q')$ \emph{unrecorded} when nothing in the output records it; a result can then change class while it still reads the same. In VaCoAl/PyVaCoAl $\rho$ is a stated setting, and the move is recorded by construction. In a learned network re-tuning can change the regime without leaving a mark in the store (\S\ref{sec:stores}). We call an unrecorded change of regime \emph{drift}. Kuhn described two states of a field, normal science and revolution. Drift is a third, in which what can be stated and how results are valued change without a crisis, without a decision by anyone who relies on it and without a record. The case is not hypothetical. When the March and June 2023 versions of two widely used language-model services were compared on the same tasks, one model's accuracy at telling prime from composite numbers fell from 84 to 51 percent, and the authors note that reproducing results from the ``same'' model becomes difficult or impossible (Chen, Zaharia \& Zou 2024). Nor is the problem confined to science: a judgement made with such a service in medicine, law or administration cannot later be traced to the regime under which it was made unless the regime was recorded. Research practice met a milder form of the problem in analytic choices made once the data are in view, and answered it by recording the choices beforehand (Nosek et al.\ 2018). The commitment of \S\ref{sec:commitment} is the corresponding record for a reading; prediction~\ref{pred:commit} states a test.

None of this calls for a single dominant paradigm to be restored. It asks only that the regime be declared and the reading recorded, so that the community has something definite to refute.

\section{Re-rooting: auditable abduction and a three-layer division of labour}\label{sec:reroot}

A rootless result can be re-rooted if it passes through three layers in turn---a generator, a walkable trace with its confidence, and a community that values it---and VaCoAl/PyVaCoAl can supply the middle layer, which is the one now missing.

\subsection{Three layers}\label{sec:layers}

The first layer generates. LLMs propose hypotheses, constructions and candidate connections fluently. Their outputs are the abductive raw material.

The second layer keeps the trace. A candidate is re-expressed as bound records and multi-hop paths in a VaCoAl/PyVaCoAl memory, where binding is exactly reversible and every retrieval carries \CRone{} and every path \CRtwo{} (2026a). This buys a trace readable constituent by constituent, a self-report faithful by construction rather than by verbalization (\href{https://arxiv.org/abs/2607.16623}{2026f}), and ranking decisions that can be reconstructed afterwards (\href{https://www.alphaxiv.org/abs/2608.17915}{2026g}). The dial is available to the user of this layer, not hidden from them: Rescue mode for exact audit, Don't Care mode for valuation, and the difference between the two read as information.

The third layer values. Which result matters, which deserves a seminar, which should be taught to the next cohort: these judgements draw on collective tacit knowledge and stay with the community (\S\ref{sec:fourpositions}). This is also where internalization and socialization happen, so it is where the tacit root actually grows, and where results are routed back, in Zoran's words, into engagement ``with what exceeds the symbolic: material resistance, bodily experience, place, living processes, and other people'' (Zoran 2026). In the terms of virtue epistemology, it is where people exercise the intellectual virtues that technology cannot take over (\S\ref{sec:virtue}). Questions raised in the third layer return to the second as new queries. That loop is how contingently tacit structure gets externalized in practice: the community asks, the substrate is given a new operation, and something that was structurally tacit becomes a stated, traceable fact.

A formal verifier, where one exists, has a definite place in this scheme, and it is not the second layer. A checker answers one question: is the result true? The second layer answers another: by what route was it reached, with what confidence at each step, and depending on what? The third answers a third: what is it worth? A verified result enters the three layers as a better candidate than an unverified one, since no effort will be spent tracing a falsehood, and it still needs a route that people can walk and a community that can value it.

\begin{figure}[tbp]
\centering
\includegraphics[width=\linewidth]{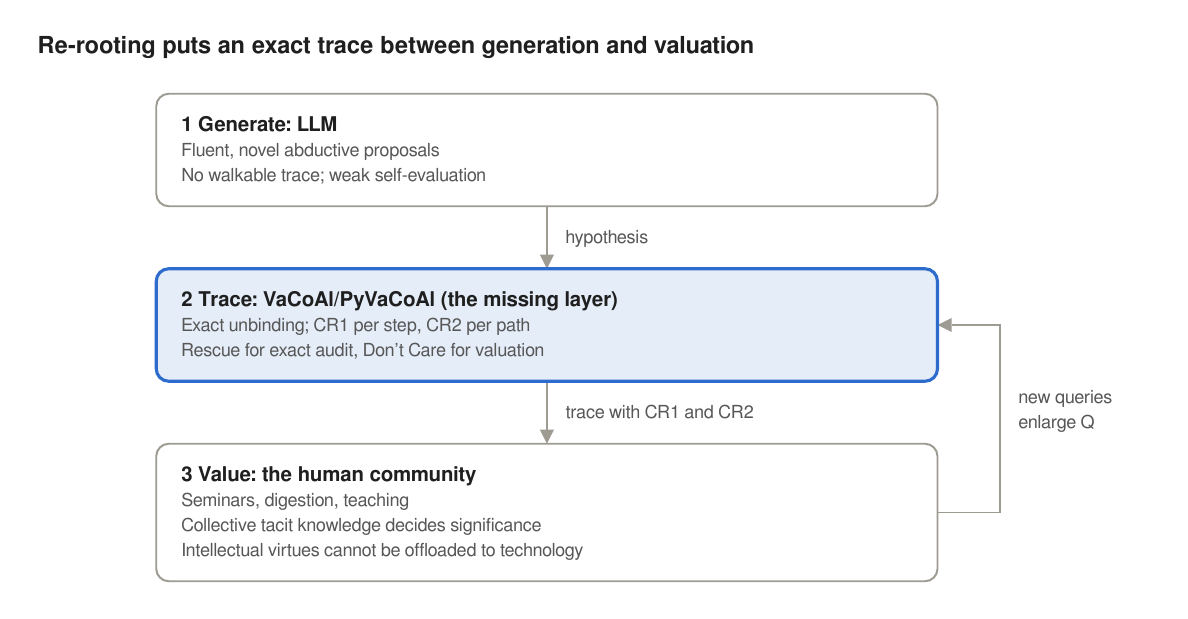}
\caption{Three layers, one feedback loop. A result moves from generation to valuation through an exact trace; questions raised by the community return to the trace layer as new queries. The third layer is where people exercise the intellectual virtues that technology cannot take over (\S\ref{sec:virtue}).}\label{fig:layers}
\end{figure}

\subsection{Why the middle layer has to be exact}\label{sec:exact}

LLMs compose facts reliably when the chain is in front of them and unreliably when it must be assembled inside the weights (Balesni, Korbak \& Evans 2024; Berglund et al.\ 2024). A multi-hop VaCoAl/PyVaCoAl traversal produces exactly what the model lacks: the intermediate nodes as stated items, in order, with a confidence for each step. This is the concrete content of the division of labour proposed in 2026a, linguistic fluency on one side and multi-hop logical integrity on the other. Exactness matters because the trace is for people. A trace that is itself approximate in an unknown way would reproduce the problem of \S\ref{sec:llm} one level down; a trace whose only approximation is a closed-form \CRtwo{} decay can be read with its error bars attached.

\subsection{What the second layer records: a commitment, not a translation}\label{sec:commitment}

An objection arises at the step from the first layer to the second. A hypothesis proposed in natural language depends on its context and can be read in more than one way; a bound record is discrete. Re-expressing the one as the other, the objection runs, must lose meaning, so the second layer inherits a translation problem at least as hard as the one it was meant to solve.

The objection assumes a fixed meaning to be preserved, and behind it a fixed truth against which a translation could be scored. Neither is given in advance. Which reading of a hypothesis is the one worth pursuing, and which of its consequences deserves a test, are no more settled beforehand than the path of a lineage is settled before it is taken. Polanyi built this openness into his idea of the real. To hold that something is real is to expect it to go on showing itself in ways not yet thought of (Polanyi 1966, 32\nobreakdash--33), and what a discoverer commits to is a bearing on such a reality, not a finished statement about it (Polanyi 1958, ch.~10).

The second layer therefore has a different task from translation. To write a hypothesis as bound records is to commit to one reading of it: these units, in these roles, along these paths. What the layer contributes is that the commitment is recorded exactly, so that it can be walked, contested and replaced. A reading that turns out to be the wrong one is then a located error and not a diffuse one. Plurality is kept at the same time: a different reading of the same proposal is one more set of records beside the first. The exactness asked for in \S\ref{sec:exact} is exactness about what was committed to. It is not a claim that the commitment exhausts what the proposal meant. Read with \S\ref{sec:falsification}, the commitment does for a reading what preregistration does for an analysis (Nosek et al.\ 2018): it fixes in advance what would count as having been wrong, so that a later change of reading is a step on the record and not a silent revision.

\subsection{The schema is a commitment too}\label{sec:schema}

The commitment begins before the reading. A reading is written in a vocabulary: the units that can be nodes, the roles that can bind them, the relations that can be edges. In the terms of \S\ref{sec:def} it fixes which queries $Q$ can be posed, and with them what can be explicit at all. It is chosen before any hypothesis is encoded, and it is not neutral: it shapes what a memory can see as theory shapes observation, and it is in this sense a paradigm in small. A post-hoc framing that the commitment keeps out of the reading can therefore come back through the schema.

A schema cannot be made neutral, but it can be declared, compared and changed on the record. Three practices follow. \emph{Plural encoders.} The same material can be encoded under schemas written independently, and the places where the encodings disagree are read as information rather than noise. If each schema is applied by more than one encoder, disagreement within a schema measures what the material leaves open, and the excess of disagreement between schemas over it measures what the schemas themselves impose. We conjecture that this excess is an operational counterpart of what Kuhn called incommensurability (Kuhn 1970, chs.~9 and 12). \emph{Public change.} A change of schema is proposed, announced and opened to comment before it takes effect. \emph{Ownership by the third layer.} What a schema should contain is a judgement about which distinctions matter, which is the work of the community; the second layer records the schema and its changes exactly. A change of schema made in public is a revolution that can be seen: the old vocabulary is kept, the new one is set beside it, and the records written in each can be compared. A change of schema made without a record is drift (\S\ref{sec:falsification}).

Who writes the records is a practical question that the proposal does not settle, and we state the division we have in mind. A generator proposes several candidate encodings of a proposal under each declared schema; the substrate checks that each is well formed and shows how it binds to the records already held; and a person chooses which to commit, working from the differences between candidates rather than from a blank page. Encoders drawn from different model families, with at least one human encoder for each schema, limit the bias they share. For a result as dense as the Jacobian counterexample the commitment may cost about as much as a digestion. What the substrate adds is that, once made, it is recorded, can be walked by others and need not be made again.

\subsection{Why the third layer is human: a virtue-epistemic reading}\label{sec:virtue}

Virtue epistemology explains knowledge as a cognitive success creditable to the knower's cognitive abilities, whether innate faculties such as perception, memory and reasoning, or acquired intellectual virtues such as conscientiousness, which involve a desire for truth (Pritchard 2023, 54\nobreakdash--59). Pritchard's discussion of technology draws a contrast that fits the first two layers. Using a calculator puts an intellectual distance between a person and the answer: the device acts as an insentient ``testifier'', and how the answer was reached may stay hidden (Pritchard 2023, 187\nobreakdash--188). An LLM's result arrives in this way. A VaCoAl/PyVaCoAl trace shortens the distance: a reader can follow the route step by step, with \CRone{} at each step and \CRtwo{} for the whole path, and so check the derivation rather than take the answer on trust.

The third layer supplies what the trace cannot. Intellectual virtues such as intellectual humility, conscientiousness and intellectual tenacity require motivation, effort and reflection, so they cannot be offloaded onto technology, however much else can be (Pritchard 2023, 189\nobreakdash--191). They are, moreover, what lead people to seek understanding rather than rest content with knowing (Pritchard 2023, 197\nobreakdash--200). Seminars, refereeing, digestion and teaching are where these virtues are exercised, and since people examine other people's arguments more strictly than their own (\S\ref{sec:dualprocess}), a trace presented in a seminar meets a stricter standard than the same trace consulted in private. The substrate can shorten the distance between a person and a result; only the person can close it.

\subsection{Designing against convergence}\label{sec:design}

Zoran ends with four conditions for those who build generative systems: ``design against convergence, preserve friction and uncertainty where they serve exploration, make provenance legible, and evaluate systems by the conditions sketched earlier in this section rather than by output quality alone'' (Zoran 2026). He derives them from art. They read, almost line by line, as a specification of the second layer (Table~\ref{tab:design}). Friction is not imperfection for its own sake. Lemma~\ref{lem:novaluation} says that a valuation exists only where perfect repair is relaxed; Zoran observes that what an engineering account treats as imperfection often functions as a record of participation. The first is a lemma about one substrate, the second an observation about practices, and what they share is a warning against optimizing a process into closure.

\begin{table}[tbp]
\centering\small
\caption{Zoran's four conditions for generative systems and their counterparts in the three-layer design.}\label{tab:design}
\smallskip
\begin{tabularx}{\linewidth}{P{0.2\linewidth}LL}
\toprule
Zoran's condition & Why it matters & Counterpart in the three-layer design \\
\midrule
Design against convergence & Closed loops of generation lose rare variations first: models trained on generated content lose the tails of the original distribution (Shumailov et al.\ 2024) & Exact records are kept rather than overwritten by generated summaries; different configurations of the substrate yield different, historically defensible valuations (Leibniz- or Gauss-dominant lineages; 2026d, Appendix~F), so plural readings can be shown instead of one converged ranking \\
Preserve friction and uncertainty where they serve exploration & Friction is where a person goes through the route; uncertainty marks where judgement is needed & \CRone{} at each step and \CRtwo{} for each path are displayed, not hidden; the difference between Rescue and Don't Care is read as information (\S\ref{sec:layers}) \\
Make provenance legible & A form whose route cannot be read cannot be extended or taught (condition~2 of \S\ref{sec:rootless-def}) & Exactly reversible binding: every item arrives with its derivation (\S\ref{sec:externalization}), and every ranking decision can be reconstructed (2026g) \\
Evaluate by these conditions, not by output quality alone & What counts is the becoming that a system makes possible & Prediction~\ref{pred:trace} measures follow-up capacity and unaided skill, not output quality \\
\bottomrule
\end{tabularx}
\end{table}

\subsection{Auditable abduction}\label{sec:abduction}

Pearl's procedure for counterfactuals begins with abduction: infer the exogenous variables consistent with the evidence, then intervene by setting a variable and cutting the arrows into it, then predict the outcome in the modified model (Pearl 2009). The hippocampal paper argues that reversible $\mathrm{GF}(2)$ binding supplies the surgical algebra of the do-operator (\href{https://arxiv.org/abs/2605.15652}{2026b}). Combined with the three layers, this gives a design we call \emph{auditable abduction}. An LLM proposes a hypothesis in Peirce's sense. The hypothesis is written into the DAG memory as bound records. Its consequences, and the counterfactual ``what if this edge were absent?'', are computed by unbind--rebind, and the change shows up as a change in CR along named paths. The human reading the result sees which bindings the hypothesis depends on before deciding what it is worth. This is a proposal, not a demonstration; the Minsky paper states that the reachability of its Reflective layer is argued rather than shown (2026f), and auditable abduction inherits that limit.

\subsection{Re-rooting already happens by hand}\label{sec:byhand}

Two episodes from 2026 show re-rooting done without such a substrate. In April, Liam Price submitted a solution to Erd\H{o}s Problem 1196 that came from his interaction with GPT-5.4 Pro; others interpreted the AI's approach and rewrote it in human-readable form, and in May a paper by Alexeev, Barreto, Li, Lichtman, Price, Shah, Tang and Tao expanded the key idea and gave a cleaner proof of the Erd\H{o}s primitive set conjecture (\href{https://terrytao.wordpress.com/2026/09/18/if-math-is-more-than-proof-we-need-to-better-celebrate-the-rest-of-it/}{Sanderson 2026}). In July, Tao's digestion rebuilt the Jacobian counterexample from multiplication of low-degree polynomials (\href{https://terrytao.wordpress.com/2026/07/21/a-digestion-of-the-jacobian-conjecture-counterexample/}{Tao 2026b}). In both episodes people did the work of the second layer by hand: they reconstructed a walkable trace before the third layer could value the result. Our claim is modest. An exact-trace substrate could make part of that reconstruction mechanical and checkable, and so leave more of the scarce human effort for understanding and teaching. It cannot replace the understanding it serves.

\section{Predictions, limitations and open problems}\label{sec:predictions}

The account makes eight predictions that could fail, and it rests on limitations a reader should weigh before accepting it.

\subsection{Predictions}\label{sec:pred}

\begin{enumerate}
\item\label{pred:rescue} \textbf{Valuation falls as rescue rises.} On the existing 25.5-million-path corpus at fixed memory depth, the dispersion of \CRtwo{} among candidates inside the frontier should fall monotonically as RR rises from 0 to 1, and the \CRtwo-ordered top-$k$ should converge on the dictionary's lexicographic top-$k$. This is Conjecture~\ref{conj:monotone}; a non-monotone curve would show that the dial is not one-dimensional and that the conjecture is false. If the dial can also serve as the exploration control of \S\ref{sec:kuhnshift}, there should be a range of intermediate rescue rates in which the number of distinct lineages in the top-$k$ rises above its value at $\mathrm{RR}=0$ before the top-$k$ turns lexicographic. Ties must not do the work: ties met while filling the frontier are broken at random over repeated draws, a lineage counts only if its \CRtwo{} lies strictly above the $(k+1)$-th value, and the count must exceed that of $k$ candidates drawn at random from the same frontier, since a rise in diversity that random sampling also produces is not exploration. If there is no such range, the dial cannot serve that role.
\item\label{pred:sort} \textbf{A test that sorts discoveries.} Any structure exposed by a new composite operation can be re-run in Rescue mode. Structure that survives is contingently tacit, an externalization in Nonaka's sense; structure that disappears is regime-bound. The coarse findings of 2026d already pass (Appendix J); the prediction is that the \CRtwo-dependent ranking results do not, and that no new discovery will fall between the two classes. A second comparison separates the substrate's valuation from a stated one: on the same frontier, rank candidates by \CRtwo{} in Don't Care mode and by hop count, the order that any attached weight $\gamma^{n}$ gives under full rescue. At fixed $N$, where the two orders differ, the difference is the part of the valuation that no stated preference for shorter routes reproduces, and whether it carries historical content is a separate question; if they never differ, the valuation of that traversal is statable (\S\ref{sec:law}).
\item\label{pred:selfread} \textbf{Self-reads fail for structural reasons.} Following 2026f, the error of a reflexive unbind of a deliberative bundle should grow with collision saturation and not with training-data coverage, the opposite signature from an LLM's self-report.
\item\label{pred:trace} \textbf{Traces re-root results in people.} Give researchers an AI-generated result in one of three forms: the result alone, the result with an LLM's self-explanation, or the result with a VaCoAl/PyVaCoAl trace and its CR values. After a fixed time, measure what Tao's follow-up activities require: answering questions on the result, presenting it, and extending it. We predict that the trace condition beats both others on questions and extension, while the self-explanation condition may score high on felt understanding without matching gains in performance. Two corollaries follow. Practitioners given AI assistance with a trace that must be walked before an output is accepted should lose less unaided skill than those given outputs alone, measured as Budzy\'{n} et al.\ (2025) and Shen and Tamkin (2026) measured it. And among formally verified AI results, matched for field and prominence, releases with a human-readable route and a record of the cull should be followed by more questions answered, talks given and variants of the result than formal proofs alone. The prediction fails if the trace makes no difference, which would mean that walking a route is not what re-roots a result.
\item\label{pred:wall} \textbf{An AI wall would be measurable.} Using a 21st-century analogue of the prosopographic graph---mentorship, co-authorship and acknowledged collaboration---results whose provenance is mainly AI-generated should show thinner downstream traffic than comparable human results within a fixed horizon, measured by the same paths-per-node statistic that separates 53.4 downstream from 5.3 upstream at Leibniz (2026d). The prediction fails if digestion of the kind in \S\ref{sec:byhand} closes the gap, which would itself be evidence that re-rooting works.
\item\label{pred:tuning} \textbf{Two tunings, one store.} Take language models of comparable scale whose documented post-training differs in kind, one shaped mainly by exact tests and one mainly by human preference. With the relevant facts in context, the first should do better on tasks that have a checker and the second on tasks judged by readers. With the context removed, so that directed relations have to be recalled from the weights alone, we predict reversal and two-hop failures of similar size in both (Berglund et al.\ 2024; Balesni, Korbak \& Evans 2024). The prediction fails if test-shaped training removes those failures, which would show that tuning changes how the store is written and not only how it is read (\S\ref{sec:stores}).
\item\label{pred:commit} \textbf{Recorded commitments locate reversals.} Take AI-generated hypotheses that the generating system later revises after a change of regime, such as re-tuning or a new rater model, and give them to two groups. One group sees the outputs before and after; the other also sees the committed readings of \S\ref{sec:commitment} for both versions. We predict that the second group locates what changed more often and more quickly, and judges more reliably whether a revision anticipates new evidence or only accommodates it after the event. In a graph whose edges are typed as observations, citations and generated outputs, seeded errors should likewise be retracted more accurately, with fewer sound edges withdrawn, when types and justifications are recorded, and fewer still when \CRtwo{} orders retraction within each type than when ties within a type are broken at random, provided that the seeded errors lie off the core; with errors seeded in the core, ordering by \CRtwo{} should do worse than random, and the ledger should single out the seeded core record as the one most often protected by adjustments that predict nothing new. And where the same hypotheses are encoded under independently written schemas (\S\ref{sec:schema}), items with a high excess of disagreement between schemas should be where conceptual divergence later appears, and items with high disagreement within schemas where clarifications appear; if the excess predicts divergence no better than disagreement within schemas does, it is not a counterpart of incommensurability. The prediction fails if the records make no difference, which would count against the claims of \S\ref{sec:falsification} that a declared regime is what lets a change of regime be judged and that recorded types and justifications are what localize retraction.
\item\label{pred:drift} \textbf{Drift can be seen from outside.} Run a fixed family of queries $Q$ against successive versions of a deployed model, as Chen, Zaharia and Zou (2024) did for accuracy on fixed tasks, and record which items each version states and which answers change. We predict changes between versions that the published notes on those versions do not mention: items that are no longer stated, and answers that change while no stated item has changed. In VaCoAl/PyVaCoAl the same probe should find such changes only where $\rho$ was changed and declared. The first part fails if every change matches a documented one; the second fails if the substrate changes class without a change of $\rho$.
\end{enumerate}

\subsection{Limitations}\label{sec:limits}

\begin{itemize}
\item \textbf{Scope and evidence.} ``Tacit'' is used in the structural sense of \S\ref{sec:def}; nothing here addresses whether a substrate can be a knower in Polanyi's personal sense. The empirical anchors come from one relation (documented mentorship), one source (Wikidata, integrating the Mathematics Genealogy Project and MacTutor) and one tracer set (the 64 Fields Medalists). Wikidata over-represents Western, university-affiliated and doctoral-advisor transmission (2026d), and the mentorship channel misses texts, correspondence and translation except where node semantics recover them.
\item \textbf{Correspondences are not identities.} The pianist, the cyclist and the Identikit are matched to VaCoAl/PyVaCoAl operations by structure, not identity. The parallel of \S\ref{sec:twodualisms} holds between popular forms that the theorists themselves reject, and Mercier and Sperber's account is one position in a continuing debate. Simon, Popper, Kuhn, Lakatos and the virtue epistemologists are read for structure, and the readings are ours: the mapping onto the functions of a paradigm, the notion of drift, and the claim that falsification beyond the endpoint needs a finer ordering than can be stated are readings, not theorems.
\item \textbf{One substrate, and only its poles.} Lemma~\ref{lem:novaluation} is proved for VaCoAl/PyVaCoAl, and the trade-off is documented at $\mathrm{RR}=1$ and $\mathrm{RR}=0$; intermediate rescue rates have not been characterized, and where the collision-born valuation departs from hop count, and whether the departures carry historical content, has not been measured (\S\ref{sec:law}). That the weights of a language model are largely a mixed store, and that tuning shapes how such a store is read and not how it is written (\S\ref{sec:stores}), are hypotheses resting on indirect evidence.
\item \textbf{Systems described from outside.} The account of the verifier lineage rests on its published descriptions, and we make no claim about how any commercial assistant is trained; the observation reported in \S\ref{sec:verifier} is one person's experience. The three culls are types, compared as kinds and not by performance. CASRAM is physically realized and commercial, but no argument here depends on its performance.
\item \textbf{Proposals are not demonstrations.} Auditable abduction (\S\ref{sec:abduction}) and the Reflective layer it relies on are designs, and the cost of encoding a proposal as bound records has not been estimated (\S\ref{sec:schema}). The code for the software version of PyVaCoAl is proprietary and available for commercial and collaborative use (2026a); 2026d states that queries, configurations, the tracer set and a reference implementation are to be deposited with a DOI on acceptance, and independent replication depends on that release.
\item \textbf{The evidence on LLMs and skills is new and will age.} The findings of \S\ref{sec:llm} come from specific models and methods, and the default rootlessness of LLM outputs may weaken as models change. The skill findings of \S\ref{sec:forms} come from one observational multicentre study and one randomized experiment with a single library. The evidence of a collective tilt (Hao et al.\ 2026) is observational, and the evidence of drift (Chen, Zaharia \& Zou 2024) concerns two versions of two services.
\end{itemize}

\subsection{Open problems}\label{sec:open}

Four questions are left open. Can audit--valuation complementarity be stated as an inequality between a measure of audit exactness and a measure of valuation contrast, with the closed-form \CRtwo{} law as its equality case? What interface lets the three layers of \S\ref{sec:reroot} work together in practice, so that a researcher can move from an LLM's proposal to a VaCoAl/PyVaCoAl trace to a seminar without losing the thread? Since the second layer records a commitment and not a translation, such an interface need not preserve a meaning; it has to show which reading was committed to and let another be set beside it. Does Lemma~\ref{lem:novaluation} have a counterpart for learned networks, so that a store cannot be both mixed enough to generalize by similarity and composed enough to be taken apart exactly, or can one system hold both? And can an unrecorded change of regime in a learned network be detected from its outputs alone, or only where the regime is declared (\S\ref{sec:falsification})?

\section{Conclusion}\label{sec:conclusion}

Tacit knowing, on this account, is an operating regime of a memory rather than a kind of content stored in it. VaCoAl/PyVaCoAl makes the claim concrete because one substrate, under one dial, yields either an exactly auditable trace or an emergent valuation of paths, and not both at full strength. Read with a relational definition of tacitness, that single trade-off gives each classical position its place, and Collins's collective kind lives in the community rather than in the substrate.

The same trade-off answers a dualism from another field. The popular moral of dual-process psychology and the popular reading of Polanyi are one picture drawn twice: a split, a ranking of the explicit half over the other, and a prescription of supervision. Mercier and Sperber's reply to the first and Polanyi's reply to the second are also one. The deliberate, explicit part is something the tacit part does, and does well mainly in the company of others. In the substrate, complete supervision is the setting at which there is no valuation left to supervise.

The mentor--student DAG is a record of tacit transmission, and Newton's thin thread, Thurston's emptied field and Tao's AI-solved problems without seminars share one structure: possession without transmission. AI-generated results are rootless by default. They can be verified, but nobody yet holds them tacitly, and their derivations leave no walkable trace. We have proposed that VaCoAl/PyVaCoAl can supply the missing middle layer between generation and valuation---an exact trace with its confidence---and stated predictions by which that proposal can fail. Understanding a result and judging its worth remain with people, because the intellectual virtues they call on cannot be offloaded to technology.

The strongest machines now at work on mathematics sharpen this diagnosis and do not remove it. The lineage from AlphaGo to AlphaProof Nexus has shown how much a test and a generator are enough for, and that a test is an exact audit of an endpoint with no valuation of its own. The tacit component has moved, to the choice of problems, to learned weights and to the digestion that a certified result still awaits, and verified but rootless is the normal product. In Popper's and Kuhn's terms, the checker keeps falsification at the endpoint, while what a paradigm used to supply has moved into a cull that gives no reasons. Beyond the endpoint, falsification is to be relocated rather than given up, and it needs three things that the older vocabularies took for granted: a committed reading, which says where a failed test bears; an ordering of what to give up first, coarse where it is stated and, by Lemma~\ref{lem:novaluation}, finer only where imperfection is tolerated; and a declared regime, so that a change of regime can be told from drift. What the substrate adds is a cull that can be walked back; what it lacks is everything the verifier lineage has achieved.

The stakes are not only those of science. Records multiply faster than the capacity to walk them, and the regimes under which judgements are made in medicine, law and administration change without record, while criticism needs something definite to criticize. On this reading, what a substrate like VaCoAl/PyVaCoAl offers is less a source of truths than a part of the infrastructure that keeps criticism possible: readings that are recorded, retractions that stay on the record, regimes that are declared and schemas that change in public, together with the people who can still walk them. People will always know more than they can state. A substrate that records how a result was reached gives them something they have rarely had: a way to show the route, which is where tacit knowledge has always begun to travel.

\section*{References}
\phantomsection\addcontentsline{toc}{section}{References}

\subsection*{The VaCoAl programme}
\begin{reflist}
\item Chuma, H., Otsuka, K., \& Sato, Y. (2026a). Beyond LLMs, sparse distributed memory, and neuromorphics: A hyper-dimensional SRAM-CAM ``VaCoAl'' for ultra-high speed, ultra-low power, and low cost. arXiv:2604.11665. \url{https://arxiv.org/abs/2604.11665}
\item Chuma, H., Otsuka, K., \& Sato, Y. (2026b). Bridging silicon and the hippocampus: Algebro-deterministic memory ``VaCoAl'' as a substrate for Vector-HaSH and TEM. arXiv:2605.15652. \url{https://arxiv.org/abs/2605.15652}
\item Chuma, H., Otsuka, K., \& Sato, Y. (2026c). How to build Marcus's algebraic mind: Algebro-deterministic substrate over Galois fields. arXiv:2605.21379. \url{https://arxiv.org/abs/2605.21379}
\item Chuma, H., Otsuka, K., \& Sato, Y. (2026d). Institutions and the transmission of upper-tail human capital: Scientific lineages across a millennium. arXiv:2605.31470. \url{https://arxiv.org/html/2605.31470}
\item Chuma, H., Otsuka, K., \& Sato, Y. (2026e). How to build Marcus's algebraic mind: From Thagard's brain--mind viewpoint. arXiv:2607.16573. \url{https://arxiv.org/pdf/2607.16573}
\item Chuma, H., Otsuka, K., \& Sato, Y. (2026f). How to build Marcus's algebraic mind: From Minsky's emotion-machine viewpoint. arXiv:2607.16623. \url{https://arxiv.org/abs/2607.16623}
\item Chuma, H., Otsuka, K., \& Sato, Y. (2026g). Computational prosopography across a millennium: Mathematically oriented lineages traced from the Fields Medalists. arXiv:2608.17915. \url{https://www.alphaxiv.org/abs/2608.17915}
\item Shuhari System (n.d.). CASRAM (SRAM-based CAM). \url{https://shuharisystem.com/?page_id=1336}
\end{reflist}

\subsection*{Tacit knowledge, machines and the history of science}
\begin{reflist}
\item Collins, H. (2010). \emph{Tacit and Explicit Knowledge}. University of Chicago Press.
\item Dreyfus, H. L. (1972). \emph{What Computers Can't Do: A Critique of Artificial Reason}. Harper \& Row.
\item Kim, M. (2026). AI is rewiring how the world's best Go players think. \emph{MIT Technology Review}, 27 February 2026. \url{https://www.technologyreview.com/2026/02/27/1133624/ai-is-rewiring-how-the-worlds-best-go-players-think/}
\item Kuhn, T. S. (1970). \emph{The Structure of Scientific Revolutions} (2nd ed.; 1st ed.\ 1962). University of Chicago Press.
\item Newell, A., \& Simon, H. A. (1976). Computer science as empirical inquiry: Symbols and search. \emph{Communications of the ACM}, 19(3), 113--126. \url{https://doi.org/10.1145/360018.360022}
\item Nonaka, I., \& Takeuchi, H. (1995). \emph{The Knowledge-Creating Company}. Oxford University Press.
\item Nonaka, I., \& von Krogh, G. (2009). Tacit knowledge and knowledge conversion: Controversy and advancement in organizational knowledge creation theory. \emph{Organization Science}, 20(3), 635--652. \url{https://doi.org/10.1287/orsc.1080.0412}
\item Peirce, C. S. (1903). Lectures on pragmatism. In \emph{Collected Papers of Charles Sanders Peirce}, Vol.~5 (1934). Harvard University Press.
\item Polanyi, M. (1958). \emph{Personal Knowledge: Towards a Post-Critical Philosophy}. University of Chicago Press.
\item Polanyi, M. (1966). \emph{The Tacit Dimension}. Doubleday.
\item Polanyi, M. (1969). \emph{Knowing and Being} (M. Grene, Ed.). University of Chicago Press.
\item Pritchard, D. (2023). \emph{What is this thing called Knowledge?} (5th ed.). Routledge. \url{https://doi.org/10.4324/9781003356110}
\item Simon, H. A. (1955). A behavioral model of rational choice. \emph{Quarterly Journal of Economics}, 69(1), 99--118.
\item Simon, H. A. (1973). Does scientific discovery have a logic? \emph{Philosophy of Science}, 40(4), 471--480. \url{https://doi.org/10.1086/288559}
\item Simon, H. A. (1976). Discussion: Bradie on Polanyi on the Meno paradox. \emph{Philosophy of Science}, 43(1), 147--151. \url{https://doi.org/10.1086/288674}
\item Tsoukas, H. (2003). Do we really understand tacit knowledge? In M. Easterby-Smith \& M. A. Lyles (Eds.), \emph{The Blackwell Handbook of Organizational Learning and Knowledge Management}. Blackwell.
\end{reflist}

\subsection*{Falsification, paradigms and research practice}
\begin{reflist}
\item Huang, K., Jin, Y., Li, R., Li, M. Y., Cand\`es, E., \& Leskovec, J. (2025). Automated hypothesis validation with agentic sequential falsifications. \emph{Proceedings of the 42nd International Conference on Machine Learning}, PMLR 267, 25372--25437. arXiv:2502.09858. \url{https://arxiv.org/abs/2502.09858}
\item Kuhn, T. S. (1977). \emph{The Essential Tension: Selected Studies in Scientific Tradition and Change}. University of Chicago Press. Cited: ch.~9, The essential tension: Tradition and innovation in scientific research (first published 1959); and Objectivity, value judgment, and theory choice (pp.~320--339).
\item Lakatos, I. (1970). Falsification and the methodology of scientific research programmes. In I. Lakatos \& A. Musgrave (Eds.), \emph{Criticism and the Growth of Knowledge} (pp.~91--196). Cambridge University Press.
\item Nosek, B. A., Ebersole, C. R., DeHaven, A. C., \& Mellor, D. T. (2018). The preregistration revolution. \emph{Proceedings of the National Academy of Sciences}, 115(11), 2600--2606. \url{https://doi.org/10.1073/pnas.1708274114}
\item Popper, K. R. (1959). \emph{The Logic of Scientific Discovery}. Hutchinson. Original work published 1934 as \emph{Logik der Forschung}.
\end{reflist}

\subsection*{Revision, records and the ecology of AI-era science}
\begin{reflist}
\item Chen, L., Zaharia, M., \& Zou, J. (2024). How is ChatGPT's behavior changing over time? \emph{Harvard Data Science Review}, 6(2). \url{https://doi.org/10.1162/99608f92.5317da47}
\item Doyle, J. (1979). A truth maintenance system. \emph{Artificial Intelligence}, 12(3), 231--272.
\item G\"{a}rdenfors, P., \& Makinson, D. (1988). Revisions of knowledge systems using epistemic entrenchment. In M. Y. Vardi (Ed.), \emph{Proceedings of the Second Conference on Theoretical Aspects of Reasoning about Knowledge} (pp.~83--95). Morgan Kaufmann.
\item Hao, Q., Xu, F., Li, Y., \& Evans, J. (2026). Artificial intelligence tools expand scientists' impact but contract science's focus. \emph{Nature}. arXiv:2412.07727. \url{https://arxiv.org/abs/2412.07727}
\end{reflist}

\subsection*{Reason, art, skill and conviviality}
\begin{reflist}
\item Budzy\'{n}, K., Roma\'{n}czyk, M., Kitala, D., et al.\ (2025). Endoscopist deskilling risk after exposure to artificial intelligence in colonoscopy: A multicentre, observational study. \emph{The Lancet Gastroenterology \& Hepatology}, 10(10), 896--903. \url{https://doi.org/10.1016/S2468-1253(25)00133-5}
\item Evans, J. St.\ B. T., \& Stanovich, K. E. (2013). Dual-process theories of higher cognition: Advancing the debate. \emph{Perspectives on Psychological Science}, 8(3), 223--241. \url{https://doi.org/10.1177/1745691612460685}
\item Kahneman, D. (2011). \emph{Thinking, Fast and Slow}. Farrar, Straus and Giroux.
\item Mercier, H., \& Sperber, D. (2017). \emph{The Enigma of Reason}. Harvard University Press.
\item Schooler, J. W., \& Engstler-Schooler, T. Y. (1990). Verbal overshadowing of visual memories: Some things are better left unsaid. \emph{Cognitive Psychology}, 22(1), 36--71. \url{https://doi.org/10.1016/0010-0285(90)90003-M}
\item Zoran, A. R. (2026). More than we can make: Generative AI, artistic forms, and artistic becoming. \emph{AI \& Society}. \url{https://doi.org/10.1007/s00146-026-03379-x}
\end{reflist}

\subsection*{Hyperdimensional computing, memory and causality}
\begin{reflist}
\item Kanerva, P. (1988). \emph{Sparse Distributed Memory}. MIT Press.
\item Pearl, J. (2009). \emph{Causality: Models, Reasoning, and Inference} (2nd ed.). Cambridge University Press.
\end{reflist}

\subsection*{Language models}
\begin{reflist}
\item Anthropic (2025a). Tracing the thoughts of a large language model, 27 March 2025. \url{https://www.anthropic.com/research/tracing-thoughts-language-model}
\item Balesni, M., Korbak, T., \& Evans, O. (2024). The two-hop curse: LLMs trained on A$\rightarrow$B, B$\rightarrow$C fail to learn A$\rightarrow$C. arXiv:2411.16353. \url{https://arxiv.org/abs/2411.16353v1}
\item Berglund, L., Tong, M., Kaufmann, M., Balesni, M., Stickland, A. C., Korbak, T., \& Evans, O. (2024). The reversal curse: LLMs trained on ``A is B'' fail to learn ``B is A''. ICLR 2024. arXiv:2309.12288. \url{https://arxiv.org/abs/2309.12288}
\item Elhage, N., Hume, T., Olsson, C., et al.\ (2022). Toy models of superposition. arXiv:2209.10652. \url{https://arxiv.org/abs/2209.10652}
\item Graepel, T. (2026). Don't be fooled---LLMs don't reason. \emph{MIT Technology Review}, 2 October 2026. \url{https://www.technologyreview.com/2026/10/02/1145639/dont-be-fooled-llms-dont-reason/}
\item Sharma, M., Tong, M., Korbak, T., Duvenaud, D., Askell, A., Bowman, S. R., et al.\ (2024). Towards understanding sycophancy in language models. ICLR 2024. arXiv:2310.13548. \url{https://arxiv.org/abs/2310.13548}
\item Shen, J. H., \& Tamkin, A. (2026). How AI impacts skill formation. arXiv:2601.20245. \url{https://arxiv.org/abs/2601.20245}
\item Shumailov, I., Shumaylov, Z., Zhao, Y., Papernot, N., Anderson, R., \& Gal, Y. (2024). AI models collapse when trained on recursively generated data. \emph{Nature}, 631(8022), 755--759. \url{https://doi.org/10.1038/s41586-024-07566-y}
\item Turpin, M., Michael, J., Perez, E., \& Bowman, S. R. (2023). Language models don't always say what they think: Unfaithful explanations in chain-of-thought prompting. NeurIPS 2023. arXiv:2305.04388. \url{https://arxiv.org/abs/2305.04388}
\end{reflist}

\subsection*{Mathematics, games and proof search in the age of AI}
\begin{reflist}
\item Hubert, T., Mehta, R., Sartran, L., et al.\ (2025). Olympiad-level formal mathematical reasoning with reinforcement learning. \emph{Nature}. \url{https://doi.org/10.1038/s41586-025-09833-y}
\item Sanderson, G. (2026). If math is more than proof, we need to better celebrate the rest of it. Guest post on \emph{What's new}, 18 September 2026. \url{https://terrytao.wordpress.com/2026/09/18/if-math-is-more-than-proof-we-need-to-better-celebrate-the-rest-of-it/}
\item Silver, D., Huang, A., Maddison, C. J., et al.\ (2016). Mastering the game of Go with deep neural networks and tree search. \emph{Nature}, 529(7587), 484--489. \url{https://doi.org/10.1038/nature16961}
\item Tao, T. (2026a). Thread on AI tools and the follow-up activities of mathematical results. Mathstodon. \url{https://mathstodon.xyz/@tao}
\item Tao, T. (2026b). A digestion of the Jacobian conjecture counterexample. \emph{What's new}, 21 July 2026. \url{https://terrytao.wordpress.com/2026/07/21/a-digestion-of-the-jacobian-conjecture-counterexample/}
\item Thurston, W. P. (1994). On proof and progress in mathematics. \emph{Bulletin of the AMS}, 30, 161--177. arXiv:math/9404236. \url{https://arxiv.org/abs/math/9404236}
\item Trinh, T. H., Wu, Y., Le, Q. V., He, H., \& Luong, T. (2024). Solving olympiad geometry without human demonstrations. \emph{Nature}, 625(7995), 476--482. \url{https://doi.org/10.1038/s41586-023-06747-5}
\item Tsoukalas, G., Kovsharov, A., Shirobokov, S., et al.\ (2026). Advancing mathematics research with AI-driven formal proof search. arXiv:2605.22763. \url{https://arxiv.org/abs/2605.22763}
\end{reflist}

\clearpage
\section*{Acknowledgments}
\phantomsection\addcontentsline{toc}{section}{Acknowledgments}

We thank the Wikidata, Mathematics Genealogy Project and MacTutor communities, whose data underlie the results reused here. The hypotheses advanced in this paper mainly originate with the first author (H.C.), grounded in the results of the authors' seven VaCoAl papers (2026a--g). Claude, an AI assistant developed by Anthropic, was used as a writing aid.

\end{document}